\documentclass[twoside]{article}

\usepackage[preprint]{aistats2027}
\usepackage[round]{natbib}
\usepackage{amsmath,amssymb}
\usepackage{graphicx}
\usepackage{tikz}
\usetikzlibrary{arrows.meta}
\usepackage{xcolor}
\usepackage{booktabs}
\usepackage{tabularx}
\usepackage{subcaption}
\usepackage{algorithm}
\usepackage{algorithmic}
\usepackage{enumitem}
\usepackage{microtype}
\usepackage{hyperref}
\usepackage{url}

\renewcommand{\bibsection}{\subsubsection*{References}}
\begin{document}

\twocolumn[
\aistatstitle{CaLA: Causal Latent Adjustment for World Models under Adaptive Policies}

\aistatsauthor{Robert Ness$^{1,*}$ \And Sofiane Ennadir$^{2,*}$ \And Oleg Smirnov$^{2,*}$}

\aistatsaddress{$^1$Microsoft Research \And $^2$King AI Labs, Microsoft Gaming}
]

\begingroup
\renewcommand{\thefootnote}{\fnsymbol{footnote}}
\footnotetext[1]{All authors contributed equally.}
\endgroup

\begin{abstract}
World models is answering ``imagine" answers to ``what-if'' questions via imagined rollouts.
But rollouts collected by adaptive policies induce treatment-outcome confounding in imagined rollouts. We study this problem in Match-3 difficulty adjustment, where latent player skill influences both assigned difficulty and churn. We build a fully specified Match-3 SCM with sequential mechanics, adaptive assignment, and paired ground-truth interventions, on which unadjusted models fit to the logged trajectories select difficulty settings that raise churn. We then introduce Causal Latent Adjustment (CaLA), a DAG-constrained world model that infers a variational player representation from pretreatment history. Across our evaluations, CaLA provides the most reliable population-level intervention estimates and competitive individualized decisions, although rich handcrafted histories can remain advantageous for individual-level adjustment. We further find that residual correction can degrade causal decision quality despite improving predictive fit, highlighting the need to distinguish predictive accuracy, confounding adjustment and the behavior of downstream causal estimators.

\end{abstract}

\section{Introduction}\label{sec:introduction}

World models support decisions by simulating environment rollouts under candidate actions~\citep{ha2018}.
However, historical policies often select logged actions using unmeasured factors.
These \emph{confounders} create noncausal associations between actions and outcomes, so a predictive model can recommend a suboptimal action.
This motivates a causal inference approach: for action $E$ and reward $R$, the causal target is $\operatorname*{argmax}_e \mathbb{E}[R \mid do(E=e)]$, which, unlike the observational $\operatorname*{argmax}_e \mathbb{E}[R \mid E=e]$, accounts for confounding and reflects the true causal consequences of an action.

Dynamic difficulty adjustment (DDA) in Match-3 makes this problem concrete~\citep{xue2017}.
Players may churn when difficulty is too high or low, so DDA policies assign difficulty from estimated skill.
Skill also affects churn directly, creating the noncausal difficulty$\leftarrow$skill$\rightarrow$churn path alongside the causal difficulty$\rightarrow$churn effect.
If skill were observed, a model could adjust for it~\citep{rezende2020}; in practice, it is latent and reflected across player history.
Better next-step prediction does not remove this bias.

We propose \textit{Causal Latent Adjustment (CaLA)}, a DAG-constrained world model that infers from pretreatment history a latent state representing common causes of historical difficulty and churn~\citep{kingma2014vae}.
CaLA holds this state fixed during rollouts and uses it for backdoor adjustment of $P(R \mid do(E=e))$~\citep{pearl2009}.
The latent state need not itself be statistically identified for the intervention distribution to be identified.
We evaluate three estimators: doubly robust direct adjustment (Direct-DR)~\citep{kennedy2017}, world-model Monte Carlo rollouts (WM-MC), and the same residual correction applied to rollouts (WM-DR).

Observational and interventional curves can differ yet share an optimum, hiding decision error.
We therefore built a playable Match-3 environment with special tiles, chain reactions, heterogeneous player skills, and DDA policies.
Its ground-truth structural causal model (SCM) creates skill-induced confounding strong enough to change the selected difficulty.
In simulated logs, CaLA better recovers ground-truth churn curves and, through world-model rollouts, reduces excess churn from player-specific recommendations by nearly 80\% relative to ignoring player history.

We contribute (1) a controlled longitudinal Match-3 game with player heterogeneity and a ground-truth SCM; (2) CaLA, which infers a static pretreatment state for a continuous treatment and holds it fixed in DAG-constrained rollouts; and (3) an evaluation framework combining natural and randomized assignment with learned and oracle player states.
The randomized control and oracle state separate latent-adjustment errors from rollout and estimator errors, while population curves and regret distinguish estimation from decision quality.

\section{Related Work}\label{sec:background}

World models answer action ``what if'' questions~\citep{ha2018}, but causal answers---especially beyond logged data---require assumptions~\citep{bareinboim2022}.
Causal partial models condition on an observed agent variable that blocks state--action confounding under policy changes~\citep{rezende2020}.
Counterfactually-Guided Policy Search uses a specified SCM to infer and compare counterfactual and logged episodes~\citep{buesing2019}.
These approaches require an observed adjustment variable or an SCM available at inference.
CaLA instead imputes a latent adjustment state with a VAE and uses the ground-truth SCM only for evaluation.

Causal effect estimators also address policy-induced confounding.
Representation methods balance observed covariates~\citep{shalit2017}, whereas CEVAE infers a latent confounder from proxies~\citep{louizos2017}.
Longitudinal models estimate outcomes under treatment sequences from observed histories~\citep{bica2020crn,melnychuk2022}, while the Time Series Deconfounder infers time-varying substitute confounders from simultaneous treatments~\citep{bica2020deconfounder}.
CaLA instead targets one continuous landmark treatment and infers a static adjustment state from repeated pretreatment history, excluding post-treatment information while conditioning direct estimators and rollouts.
It combines generalized propensity scores~\citep{hirano2004}, cross-fitted nonparametric doubly robust estimation~\citep{kennedy2017,chernozhukov2018}, and a propensity-weighted residual correction related to doubly robust off-policy evaluation~\citep{jiang2016}.

DDA adapts challenge from player state or behavior~\citep{xue2017}.
Prior Match-3 work estimates difficulty from player data~\citep{kristensen2021} or AI players~\citep{roohi2021}, but does not separate errors in causal adjustment, simulated dynamics, and recommendations.
Our playable Match-3 SCM supplies sequential mechanics, latent heterogeneity, adaptive assignment, and paired interventional outcomes to evaluate these errors separately.

\section{A Controlled Match-3 Structural Causal Model}\label{sec:match3-scm}

Because logged gameplay alone cannot reveal what a different difficulty would have caused, we construct a playable Match-3 SCM. Our environment is inspired by existing games and captures realistic game mechanics and stylized player dynamics while enabling full experimental control over every structural mechanism.

\subsection{Environment Mechanics and Structural Equations}
Each player $i$ has a fixed latent state $K_i \in \mathbb{R}^4$ whose components represent search ability, pattern recognition, planning horizon, and strategic execution. Only the simulator and the oracle diagnostics built on it can access $K_i$. Through this state, a player's skills affect the difficulty assigned to them, the actions they take, and their retention. The board dynamics do not depend on it. Appendix~\ref{app:scm-graphs} depicts the full causal graphs.

Attempt $j$ of player $i$ is described by a level context $L_{ij}$ and a baseline tier $D_{ij}$. The context sets the color distribution, refill weighting, and feature demands, whereas the tier fixes the move budget. The historical adaptive policy then draws a continuous difficulty
\begin{equation}
E_{ij} \sim \mathcal N \left(
 b(D_{ij},L_{ij})+\gamma_{L_{ij}}z_{iL_{ij}}, \sigma_{L_{ij}}^2 \right),
\label{eq:difficulty-assignment}
\end{equation}
with $z_{il}$ denoting the standardized projection of $K_i$ onto the demands of level $l$. A positive $\gamma_l$ assigns higher difficulty to more skilled players, which induces policy confounding.

A board state $S_{ijt}$ consists of parallel $8 \times 8$ color and special-tile grids, together with goal progress, move counters, and a terminal flag. Each attempt opens from an initial board $S_{ij0} \sim p_0(\,\cdot \mid L_{ij}, D_{ij}, E_{ij})$, with the goal quota set by the served difficulty. Figure~\ref{fig:heldout-summary} (left) shows a representative opening state rendered by the simulator. An action $A_{ijt}$ swaps two adjacent cells and must be legal on the current board. States evolve according to $S_{ij,t+1} \sim p(\,\cdot \mid S_{ijt}, A_{ijt}, L_{ij}, D_{ij}, E_{ij})$, so that $K_i$ can affect a state-action trajectory only through the choice of $A_{ijt}$, while matching, cascades, gravity, and tile refills work the same way for every player. Appendix~\ref{app:engine-details} provides further engine details, and Figure~\ref{fig:gameplay-examples} illustrates representative cascade and striped-tile mechanics.

When an attempt reaches its terminal state, it produces a binary completion $R_{ij} \in \{0, 1\}$ and a signed margin $Q_{ij} \in [-1, 1]$, which measures the margin of victory or defeat. After every attempt, a recency-weighted experience variable is updated as $M_{i,j+1} = M_{ij} + \rho (R_{ij} - M_{ij})$, with the calibrated update rate $\rho = 0.554$ and the shared initial value $M_{i1} = 0.339$. Churn on the next attempt, $C_{i,j+1} \in \{0, 1\}$, is our retention outcome. It is modeled as an absorbing Bernoulli event whose hazard depends on the deviations of $M_{i,j+1}$ and $Q_{ij}$ from the level's targets. Being over-challenged raises the hazard more steeply than being under-challenged, so the churn-minimizing difficulty lies in the interior of the range. This shape is consistent with earlier work on the relationship between challenge and retention~\citep{reguera2020quantifying}, although recent experimental studies also report engagement that is insensitive to difficulty-skill balance~\citep{cutting2023}. The benchmark only requires an interior optimum that makes the decision problem nontrivial, not fidelity to human churn.

\subsection{Interventional Queries and Ground-Truth Evaluation}
We assess the causal decision at the landmark attempt $j=20$, which isolates a single continuous intervention after a fixed pretreatment history while retaining sequential within-attempt dynamics. Our analysis therefore concerns a landmark intervention rather than identification or optimization of a fully time-varying treatment policy. For a player who is still active at that point ($C_{i,20}=0$) and plays level $l$, let $C_{i,21}$ denote churn before attempt $21$. The observational response curve describes the players to whom the logging policy happened to assign difficulty $e$, whereas the interventional curve fixes the difficulty for everyone in the target population:
\begin{equation}
\begin{aligned}
m_{\mathrm{obs},l}(e) &= P\big(C_{i,21}=1 \mid E_{i,20}=e, \mathcal{V}_l\big), \\
m_{\mathrm{do},l}(e) &= P\big(C_{i,21}=1 \mid do(E_{i,20}=e), \mathcal{V}_l\big),
\end{aligned}
\label{eq:response-curves}
\end{equation}
where $\mathcal{V}_l = \{L_{i,20}=l,\, C_{i,20}=0\}$ denotes the active cohort on level $l$.
Minimizing the two gives the observational choice $\widetilde{e}_l = \arg\min_e m_{\mathrm{obs},l}(e)$ and the causal optimum $e_l^* = \arg\min_e m_{\mathrm{do},l}(e)$.

To obtain ground truth interventional curves from the structural mechanisms, rather than from a learned outcome model, we fix each player's pre-intervention history and exogenous randomness. For each candidate difficulty $e$, we replace the historical assignment mechanism with the intervention $do(E=e)$ and simulate gameplay using the true behavioral and environment mechanisms. Reusing the same player attributes and exogenous noise across candidate difficulties yields paired counterfactual outcomes whose differences are attributable to the intervention, following the standard abduction--action--prediction construction~\citep{pearl2009,buesing2019}.

\subsection{Empirical Demonstration of Confounding Bias}

\begin{figure*}[t]
\centering
\includegraphics[width=0.18\textwidth]{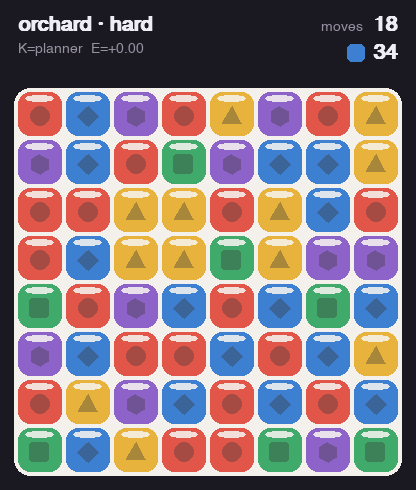}
\hfill
\includegraphics[width=0.76\textwidth]{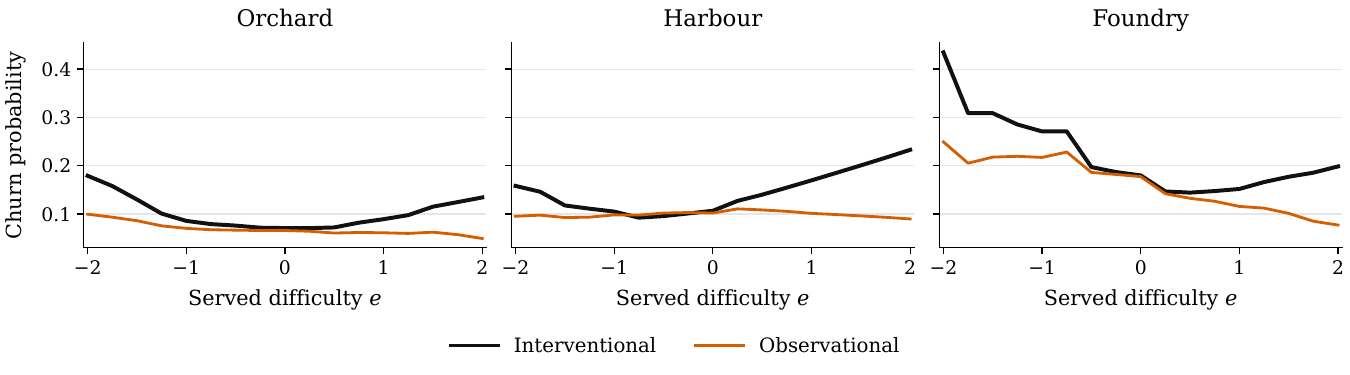}
\caption{Controlled Match-3 benchmark. Left: simulator-rendered opening state for the Orchard level at fixed difficulty $E=0$; the HUD shows the move budget and goal quota. Right: observational and interventional response curves across the three level families.}
\label{fig:heldout-summary}
\end{figure*}

We verify that the gap appears in practice by comparing standard logging-policy estimates with paired counterfactual simulations on held-out cohorts across the three level families. In every scenario, the observational models fit to the logged trajectories select the hardest candidate difficulty ($\widetilde{e}_l = 2.00$). Figure~\ref{fig:heldout-summary} depicts the resulting observational and interventional response curves, while Table~\ref{tab:heldout-summary} shows that the ground-truth causal optimum $e_l^*$ varies across the interior of the candidate grid. Across seed and level combinations, selecting difficulty from the observational curve increases churn by approximately $5.3$ to $12.9$ percentage points relative to the interventional optimum.

\begin{table*}[t]
\caption{Observational and interventional difficulty recommendations.
Ranges span five benchmark seeds.}
\label{tab:heldout-summary}
\centering
\scriptsize
\begin{tabular}{@{}lcccc@{}}
\toprule
\textbf{Level}
& \shortstack{\textbf{Observational}\\\textbf{recommendation}}
& \shortstack{\textbf{Interventional}\\\textbf{optimum}}
& \shortstack{\textbf{Recommendation}\\\textbf{gap} $\downarrow$}
& \textbf{Added churn} $\downarrow$ \\
\midrule
Orchard & $+2.00$ & $-0.25$ to $+0.25$
& $1.75$--$2.25$ & $0.053$--$0.069$ \\
Harbour & $+2.00$ & $-0.75$
& $2.75$ & $0.109$--$0.129$ \\
Foundry & $+2.00$ & $+0.25$ to $+0.50$
& $1.50$--$1.75$ & $0.079$--$0.088$ \\
\bottomrule
\end{tabular}
\end{table*}

\section{Causal Latent Adjustment} \label{sec:cala}

\textit{Causal Latent Adjustment (CaLA)} combines an SCM with a variational world model to simulate difficulty interventions while explicitly encoding assumptions about confounding adjustment and mechanism separation. It targets the policy-induced confounding established in Section~\ref{sec:match3-scm}.

\subsection{Problem Formulation and Causal Targets}
For the landmark decision defined in Section~\ref{sec:match3-scm}, the served difficulty $E_i \in \mathbb{R}$ is the continuous treatment, and next-attempt churn $Y_i = C_{i,21} \in \{0,1\}$ is the outcome of interest. The level context $L_i$ gives the task family, while the decision-time context $B_i$ collects the observable baseline properties, namely the level, tier, and reconstructed prior experience $M_{i,20}$. The pretreatment history is $H_{i,<20} = (X_{i0}, O_{i1}, O_{i2}, \dots, O_{i,19})$, where $X_{i0}$ is the optional baseline assessment collected before the first attempt and each $O_{ir}$ with $r<20$ contains the task context, assigned difficulty, behavioral summaries, game outcomes, and proxy measurements for that attempt. Attempt 20 and everything downstream are excluded to avoid post-treatment bias. For active players on level $l$, we consider two interventional quantities.

The population intervention curve is the expected outcome over the active cohort when difficulty is set by intervention:
\begin{equation}
    m_l(e) = \mathbb{E}\big[Y_i(e) \mid L_i = l, C_{i,20} = 0\big].
\label{eq:population-intervention}
\end{equation}

The individual conditional curve is the same quantity for a particular player, conditioned on that player's pretreatment history:
\begin{equation}
r_i(e)
= \mathbb{E}\big[
Y_i(e) \mid H_{i,<20}, B_i,
L_i=l, C_{i,20}=0
\big].
\label{eq:individual-intervention}
\end{equation}
The ground-truth simulator operates on an unobserved skill vector $K_i$. CaLA does not observe $K_i$ and instead infers a continuous latent state $Z_i$ from $H_{i,<20}$ and $B_i$. The point of $Z_i$ is not to recover skill semantically but to retain common-cause information relevant to the backdoor paths from $E_i$ to $Y_i$.

\subsection{DAG-Constrained Generative Factorization}
CaLA uses the causal DAG to constrain how the generative world model factorizes. The latent state $Z_i$ enters the model as a common cause of the historical difficulty assignment $E_i$, the player's actions $A_{it}$, and the churn outcome $Y_i$, but it does not appear in the environment transition $p_\theta(S_{i,t+1} \mid S_{it}, A_{it}, L_i, D_i, E_i)$. Conditioned on the latent state $Z_i$ and the observed context $(B_i, L_i, D_i)$, the trajectory and outcome are then modeled as
\begin{equation}
\begin{aligned}
&p(E_i, S_{i,\le T_i}, A_{i,<T_i}, Y_i \mid Z_i, B_i, L_i, D_i) \\
&\quad= g_\phi(E_i \mid B_i, Z_i, L_i, D_i) \\
&\qquad\times p_\theta(S_{i0} \mid L_i, D_i, E_i) \\
&\qquad\times \prod_{t < T_i}
p_\psi(A_{it} \mid S_{it}, Z_i, L_i, D_i, E_i) \\
&\qquad\qquad\times
p_\theta(S_{i,t+1} \mid S_{it}, A_{it}, L_i, D_i, E_i) \\
&\qquad\times p_\gamma(Y_i \mid R_i, Q_i, M_{i,21}, B_i, Z_i, L_i),
\end{aligned}
\label{eq:cala-factorization}
\end{equation}
where $g_\phi$ is the propensity density, $p_\psi$ is the behavioral policy over legal moves, $p_\theta$ is the environment model, and $p_\gamma$ is the retention model. The standard normal prior over $Z_i$ completes the joint distribution. The amortized posterior $q_\eta(Z_i \mid H_{i,<20}, B_i)$ of~\eqref{eq:latent-posterior} approximates the intractable $p(Z_i \mid H_{i,<20}, B_i)$.

Leaving $Z_i$ out of $p_\theta$ is a statement about the environment rather than a restriction adopted for lack of capacity. A more skilled player chooses different actions $A_{it}$, but the game responds to those actions in the same way for everyone (Section~\ref{sec:match3-scm}). If $Z_i$ were allowed into the transition kernel, the world model could explain policy-dependent outcomes as if the dynamics themselves had changed. Excluding it rules this out.

\subsection{Inferring the Latent Adjustment State}
The encoder maps the pretreatment history to a Gaussian with diagonal covariance,
\begin{equation}
\begin{aligned}
q_\eta(Z_i \mid H_{i,<20}, B_i)
&= \mathcal{N}\Big(
    \mu_\eta(H_{i,<20}, B_i), \\
&\qquad \operatorname{diag}\{
    \sigma_\eta^2(H_{i,<20}, B_i)\}
    \Big).
\end{aligned}
\label{eq:latent-posterior}
\end{equation}
It is a recurrent network that reads attempts $r<20$ in order. From each prefix, auxiliary heads predict the historically assigned difficulty, behavioral summaries, completion, and proxy evidence. Training minimizes the resulting auxiliary prediction loss, in expectation under the posterior, together with a KL penalty that pulls the posterior toward a standard Gaussian prior:
\begin{equation}
\begin{aligned}
\mathcal{L}_{\mathrm{CaLA}}
&= \mathbb{E}_{q_\eta}\big[\mathcal{L}_{\mathrm{aux}}(Z_i)\big] \\
&\quad + \beta D_{\mathrm{KL}}\Big(
q_\eta(Z_i \mid H_{i,<20}, B_i)
\;\|\; \mathcal{N}(0, I)\Big).
\end{aligned}
\label{eq:cala-objective}
\end{equation}
At evaluation time, we average over posterior draws $Z_i^{(s)} \sim q_\eta$ rather than collapsing to the mean, so that representation uncertainty carries through to the estimates.

\subsection{Interventional Simulation and $do$-Rollouts}
Given the intervention $do(E_i = e)$, the evaluator drops the logging policy $g_\phi(E_i \mid \cdot)$, sets $E_i$ to $e$, and ancestrally samples the remaining factors of Eq.~\eqref{eq:cala-factorization} with the player's inferred $Z_i^{(s)} \sim q_\eta$ held fixed rather than resampled from the population that historically received $e$. It samples the opening board from $p_\theta$, then alternates legal actions from $p_\psi$ and transitions from $p_\theta$ until termination, computes the completion $R_i(e)$, margin $Q_i(e)$, and updated experience $M_{i,21}(e)$, and evaluates the churn risk $\widehat{\mu}_{\mathrm{WM},l}(e, B_i, Z_i^{(s)}) = p_\gamma(Y_i = 1 \mid R_i, Q_i, M_{i,21}, B_i, Z_i^{(s)}, L_i)$, averaged over $S$ posterior draws and $N_{\mathrm{roll}}$ rollout replicates (Algorithm~\ref{alg:cala-rollout}, Appendix~\ref{app:cala-implementation}). Reusing the same latent draws and random number streams across candidate difficulties pairs the evaluations for a given player and reduces the Monte Carlo variance of comparisons.

\subsection{Assumptions and Estimation}
Estimation targets the interventional curves through the following assumptions. The first five are the standard conditions under which the adjusted functional equals $m_l(e)$, while the final assumption is specific to the world-model estimators:

\begin{enumerate}[label=\arabic*)]
\item \textbf{Consistency.} The observed outcome equals the potential outcome under the treatment that was actually assigned.
\item \textbf{Strict Temporal Ordering.} The adjustment state $Z_i$ is computed from pretreatment information only ($r < 20$).
\item \textbf{Adjustment Sufficiency (Unconfoundedness).} Given $W_i = (B_i, Z_i)$, treatment is independent of the potential outcomes, $Y_i(e) \perp\perp E_i \mid W_i, L_i$.
\item \textbf{Conditional Positivity (Overlap).} The propensity density satisfies $g_l(e \mid W_i) > 0$ for every candidate difficulty we evaluate.
\item \textbf{Non-Interference.} A player's assigned difficulty has no effect on any other player's outcome.
\item \textbf{Mechanism Separation.} $Z_i$ enters the assignment policy, player behavior, and retention but does not enter the transition dynamics $p_\theta$.
\end{enumerate}

Assumption 3 is the substantive one, and it is not testable from observational data. Fitting a latent-variable model does not make it hold: proxy-based identification requires exclusion restrictions and completeness conditions~\citep{miao2018}, and variational recovery of a latent confounder is consistent only in restricted settings~\citep{rissanen2021}. We therefore treat adjustment sufficiency as a working hypothesis and use the paired engine ground truth and the randomized control to measure how far learned adjustment actually gets.

On the training cohorts, with cross-fitting~\citep{chernozhukov2018}, we fit an observational outcome regression $\widehat{\mu}_{\mathrm{obs},l}(e, w)$ and a generalized propensity density $\widehat{g}_l(e \mid w)$~\citep{hirano2004}. Three estimators of $m_l(e)$ are then built from these components.

The direct doubly robust adjustment estimator $\widehat{m}_l^{\mathrm{Direct\text{-}DR}}$ follows nonparametric doubly robust estimation for continuous treatments~\citep{kennedy2017}. The outcome regression is corrected by a propensity-weighted residual localized around $e$ with a kernel:
\begin{equation}
\begin{aligned}
\widehat{m}_l^{\mathrm{Direct\text{-}DR}}(e)
&= \frac{1}{n_l}\sum_{i:L_i=l}
\Bigg[
\widehat{\mu}_{\mathrm{obs},l}(e,W_i) \\
&\quad+
\frac{K_h(E_i-e)}
{\widetilde{g}_l(E_i\mid W_i)}
\Big\{Y_i-\widehat{\mu}_{\mathrm{obs},l}(E_i,W_i)\Big\}
\Bigg].
\end{aligned}
\label{eq:direct-dr}
\end{equation}
Here $K_h(u)=K(u/h)/h$ is a kernel with bandwidth $h$ and $\widetilde{g}_l=\max(\widehat{g}_l, g_{\min})$ is the propensity density floored at $g_{\min}$; $h$, $g_{\min}$, and the individual-support threshold are fixed before test evaluation.

World-model Monte Carlo, $\widehat{m}_l^{\mathrm{WM\text{-}MC}}$, replaces the regression with the churn risk $\widehat{\mu}_{\mathrm{WM},l}(e, W_i)$ obtained from autonomous world-model rollouts (Algorithm~\ref{alg:cala-rollout}):
\begin{equation}
\widehat{m}_l^{\mathrm{WM\text{-}MC}}(e) = \frac{1}{n_l} \sum_{i:L_i=l} \widehat{\mu}_{\mathrm{WM},l}(e, W_i).
\label{eq:wm-mc}
\end{equation}

World-model doubly robust estimation, $\widehat{m}_l^{\mathrm{WM\text{-}DR}}$, uses the rollout risk as the outcome model and applies the same residual correction:
\begin{equation}
\begin{aligned}
\widehat{m}_l^{\mathrm{WM\text{-}DR}}(e)
&= \frac{1}{n_l}\sum_{i:L_i=l}
\Bigg[
\widehat{\mu}_{\mathrm{WM},l}(e,W_i) \\
&\quad+
\frac{K_h(E_i-e)}
{\widetilde{g}_l(E_i\mid W_i)}
\Big\{Y_i-\widehat{\mu}_{\mathrm{WM},l}(E_i,W_i)\Big\}
\Bigg].
\end{aligned}
\label{eq:wm-dr}
\end{equation}

The residual correction is defined at the population level. After posterior integration, individual recommendations minimize the supported direct-outcome or WM-MC surface.

\section{Experimental Evaluation}\label{sec:experiments}
We evaluate CaLA in the controlled Match-3 environment through two research questions.

Our first research question, \emph{RQ1 (CaLA Adjustment)}, asks whether CaLA's variational encoder can infer a latent state $Z_i$ from pretreatment that captures the information relevant to confounding adjustment and improves decision quality under direct adjustment. 

Our second research question, \emph{RQ2 (CaLA Imagination)}, asks whether CaLA's DAG-constrained world model, which holds $Z_i$ fixed during autonomous interventional rollouts, can approximate ground-truth intervention curves and support low-regret difficulty selection, and how these estimates are affected by residual correction. 

We evaluate all methods on $2{,}400$ simulated player identities under two assignment regimes: \emph{natural assignment}, in which difficulty adapts to oracle player skill $K_i$, and \emph{randomized assignment}, a negative control in which skill dependence is removed while marginal treatment variance is preserved. Players are partitioned into training, validation, and test sets. Appendix~\ref{app:data-details} provides the complete data and treatment protocol.

We compare five adjustment arms that vary the pretreatment information available to the downstream estimators. Table~\ref{tab:adjustment-arms} summarizes their information sets and structural roles. We evaluate intervention curves on the 17-point grid $\mathcal{E}=\{-2.00,-1.75,\ldots,1.75,2.00\}$.

\begin{table*}[t]
\caption{Information available to the adjustment arms before the landmark treatment.}
\label{tab:adjustment-arms}
\centering
\footnotesize
\begin{tabularx}{\textwidth}{@{}lXX@{}}
\toprule
\textbf{Adjustment arm} & \textbf{Information available before attempt 20} & \textbf{Structural role} \\
\midrule
No history & Level context $L$, tier $D$, and reconstructed experience $M_{20}$ & Observational baseline \\
Handcrafted history & Prespecified rates, means, slopes, and counts from attempts 1--19 & Feature-engineering baseline \\
Deterministic $Z$ & Point encoding $f_\eta(H_{<20},B)$ & Deterministic CaLA ablation \\
Variational CaLA & Posterior $q_\eta(Z\mid H_{<20},B)$ & Proposed adjustment (CaLA encoder) \\
Oracle $K$ & True simulator skill vector $K_i$ & Evaluation-only diagnostic \\
\bottomrule
\end{tabularx}
\end{table*}

Each adjustment arm is evaluated with the three estimators introduced in Section~\ref{sec:cala}, which differ in the source of the outcome-risk estimate and in whether a residual correction is applied. Direct-DR takes its risk from the logged-outcome regression $\widehat{\mu}_{\mathrm{obs}}$ and adds the propensity-weighted residual correction; it serves as the observational anchor. WM-MC instead uses the risk $\widehat{\mu}_{\mathrm{WM}}$ obtained from autonomous world-model rollouts without residual correction, while WM-DR applies the same correction to the rollout-based risk. All three estimators are evaluated against the paired engine ground truth under $do(E=e)$.

For held-out test players, \emph{curve MAE} measures absolute error between estimated and ground-truth intervention curves. \emph{Decision regret} is the ground-truth churn under a method's recommended difficulty minus churn at the true causal optimum.

\section{Results}\label{sec:results}

\subsection{Direct Latent Adjustment (RQ1)}

Appendix Table~\ref{tab:direct-dr-aggregate} collects the aggregate Direct-DR errors and regrets, and Appendix Figure~\ref{fig:direct-adjustment-population} compares the population curves with paired engine interventions. Under natural assignment, every history-conditioned arm improves on the no-history baseline. Variational CaLA attains the lowest mean population regret and pooled individual regret, reducing the individual regret by approximately $62\%$ relative to no history. Oracle $K$ attains the lowest weighted curve MAE, while CaLA has the lowest MAE among the non-oracle arms.

The detailed natural-assignment results in Appendix Table~\ref{tab:direct-dr-natural-level} show that these population summaries conceal substantial level variation. The recommendation from CaLA's mean curve matches the engine optimum in Orchard and Foundry, while selecting an easier difficulty in Harbour. In Orchard, handcrafted history selects $+1.50$. Whole-player bootstrap intervals also show that point-optimal recommendations can remain unstable under resampling. Appendix Figure~\ref{fig:direct-adjustment-individual} shows the player-level regret distributions.

Under randomized assignment, CaLA attains the lowest displayed mean population regret, while deterministic $Z$ has the lowest curve MAE and handcrafted history the lowest pooled individual regret. Residual estimation error remains even for oracle $K$. These results support the value of pretreatment information for individualized decisions, but neither learned adjustment nor access to the true skill vector guarantees accurate recovery of the full population intervention curve. 

\subsection{World-Model Rollouts (RQ2)}

\begin{figure*}[t]
\centering
\includegraphics[width=\textwidth]{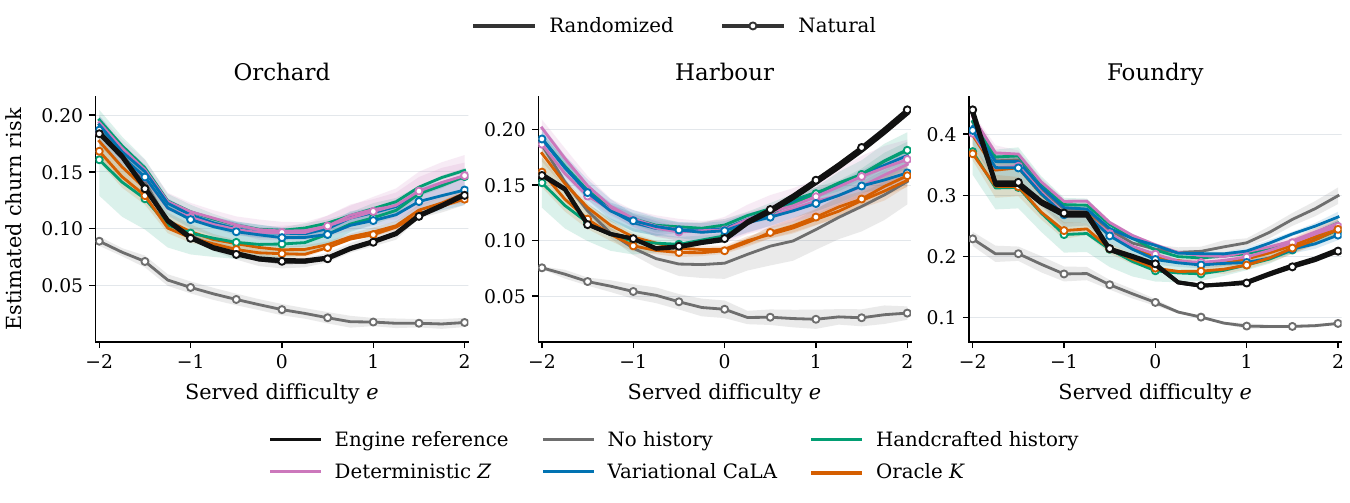}
\caption{WM-MC population intervention curves under randomized (solid) and natural (hollow markers) assignment. Black curves show the paired engine reference. Colored curves and shaded bands show means and $\pm$ one sample standard deviation across three seeds.}
\label{fig:wm-mc-population}
\end{figure*}

Table~\ref{tab:wm-aggregate} reports the aggregate WM-MC and WM-DR results while Figure~\ref{fig:wm-mc-population} provides the corresponding WM-MC response curves. Under natural assignment, CaLA achieves low weighted curve MAE and mean population regret. Among the three history-conditioned arms, CaLA has the lowest unrounded mean population regret, while handcrafted history has the lowest curve MAE. Under randomized assignment, CaLA has the lowest curve MAE and population regret within this matched history-arm comparison. The per-level estimates and recommendations appear in Appendix Table~\ref{tab:wm-mc-natural-level}.

\begin{table*}[t]
\caption{Aggregate WM-MC and WM-DR performance against engine reference. Values are means $\pm$ sample SDs across three policy-training seeds. All rows use 128 imagined trajectories per player--difficulty. Individual regret is shared by WM-MC and WM-DR.}
\label{tab:wm-aggregate}
\centering
\scriptsize
\resizebox{\textwidth}{!}{%
\begin{tabular}{llccccc}
\toprule
\textbf{Regime} & \textbf{Adjustment arm}
& \shortstack{\textbf{WM-MC Weighted}\\\textbf{Curve MAE} $\downarrow$}
& \shortstack{\textbf{WM-MC Mean}\\\textbf{Population regret} $\downarrow$}
& \shortstack{\textbf{WM-DR Weighted}\\\textbf{Curve MAE} $\downarrow$}
& \shortstack{\textbf{WM-DR Mean}\\\textbf{Population regret} $\downarrow$}
& \shortstack{\textbf{Pooled Individual}\\\textbf{Regret} $\downarrow$} \\
\midrule
Natural & No history
& $0.079 \pm 0.004$ & $0.037 \pm 0.010$
& $0.058 \pm 0.000$ & $0.060 \pm 0.000$
& $0.078 \pm 0.006$ \\
Natural & Handcrafted history
& $\mathbf{0.019 \pm 0.002}$ & $0.003 \pm 0.003$
& $0.030 \pm 0.004$ & $0.006 \pm 0.003$
& $\mathbf{0.014 \pm 0.000}$ \\
Natural & Deterministic $Z$
& $0.021 \pm 0.003$ & $0.002 \pm 0.001$
& $\mathbf{0.020 \pm 0.001}$ & $\mathbf{0.005 \pm 0.003}$
& $0.016 \pm 0.002$ \\
Natural & Variational CaLA
& $0.019 \pm 0.004$ & $\mathbf{0.002 \pm 0.000}$
& $0.023 \pm 0.001$ & $0.007 \pm 0.003$
& $0.016 \pm 0.002$ \\
Natural & Oracle $K$
& $0.014 \pm 0.001$ & $0.003 \pm 0.001$
& $0.021 \pm 0.001$ & $0.011 \pm 0.006$
& $0.014 \pm 0.001$ \\
\midrule
Randomized & No history
& $0.019 \pm 0.003$ & $0.003 \pm 0.001$
& $0.061 \pm 0.001$ & $0.021 \pm 0.000$
& $0.038 \pm 0.002$ \\
Randomized & Handcrafted history
& $0.024 \pm 0.003$ & $0.002 \pm 0.001$
& $0.074 \pm 0.001$ & $0.032 \pm 0.002$
& $\mathbf{0.013 \pm 0.001}$ \\
Randomized & Deterministic $Z$
& $0.025 \pm 0.005$ & $0.002 \pm 0.001$
& $0.067 \pm 0.002$ & $\mathbf{0.019 \pm 0.006}$
& $0.015 \pm 0.000$ \\
Randomized & Variational CaLA
& $\mathbf{0.022 \pm 0.004}$ & $\mathbf{0.002 \pm 0.000}$
& $\mathbf{0.066 \pm 0.001}$ & $0.026 \pm 0.000$
& $0.014 \pm 0.001$ \\
Randomized & Oracle $K$
& $0.016 \pm 0.001$ & $0.003 \pm 0.001$
& $0.068 \pm 0.002$ & $0.019 \pm 0.006$
& $0.013 \pm 0.002$ \\
\bottomrule
\end{tabular}%
}
\end{table*}

Appendix Figure~\ref{fig:wm-individual-regret} shows the player-level recommendation regret distributions. Relative to the displayed no-history comparator, CaLA reduces individual regret by almost $80\%$ under natural assignment and $64\%$ under randomized assignment. Handcrafted history yields further reductions of approximately $12\%$ and $7\%$, respectively, relative to CaLA.

\subsection{Diagnostics and Sensitivity}
Residual correction does not improve CaLA's population estimates in these runs. Under natural assignment, WM-DR increases weighted curve MAE by approximately $19\%$ and mean population regret by approximately $238\%$. Under randomized assignment, both metrics also increase, and curve-error deterioration occurs for every adjustment arm, including oracle $K$. WM-DR therefore serves primarily as a diagnostic of sensitivity to residual correction rather than evidence of a uniform improvement over imagination alone. Appendix Table~\ref{tab:wm-dr-population} and Figure~\ref{fig:wm-dr-appendix} provide the complete results.

The transition study in Appendix~\ref{app:world-model-architecture-and-rollout-evidence} finds that the Next-Board Transformer has the strongest predictive accuracy and rollout stability across the evaluated parameter bands. Downstream intervention performance is less monotone: Appendix Table~\ref{tab:cala-wm-capacity} shows that increasing capacity does not consistently reduce curve error or regret, and that LeWM can match or improve a Transformer on selected decision metrics despite weaker state prediction.

\section{Discussion and Conclusion}
\label{sec:discussion-conclusion}

The Match-3 experiments expose a decision failure, not merely a curve-fitting error. Specifically, under DDA-confounded assignment, observational and interventional curves can select different difficulties.
CaLA partly closes this gap.
In particular, pretreatment history contains common-cause information that improves individualized decisions, and CaLA performs strongly under confounded assignment.
Its learned state does not uniformly outperform handcrafted features on population metrics.
Even so, CaLA can also use this state when generating rollouts.

These results also show that accurate next-state prediction does not guarantee accurate estimates of what would happen if the difficulty changed.
For example, stronger transition models are not uniformly better for downstream decisions, and greater capacity does not consistently reduce regret.
World-model evaluation must therefore test whether rollouts recover the effects of changing difficulty, not only whether they predict the next state.

\subsection{Diagnostics and Limitations}

The randomized-assignment control localizes WM-DR's failure. It increases CaLA's curve error and regret under both assignment rules and still degrades estimates with randomized assignment and oracle $K$, implicating the continuous-treatment estimator rather than latent inference. We cannot distinguish finite-sample nuisance error, kernel localization, and residual correction. Thus, WM-MC is more reliable here, while WM-DR remains a log-only sensitivity diagnostic: a large correction signals disagreement between rollout predictions and propensity-weighted outcomes.

CaLA does not identify a unique confounder: training encourages $Z$ to retain assignment, behavior, and outcome information but neither proves adjustment sufficiency~\citep{damour2019} nor gives $Z$ a unique causal meaning.
Here, oracle $K$ and paired interventions test this limitation; deployment would instead require defensible proxy assumptions and sensitivity analyses.

The engine provides realistic sequential mechanics, but its player attributes, assignment rule, and churn mechanism are constructed. CaLA cannot estimate effects beyond difficulties assigned to similar histories. We report curve error with regret because component errors can cancel at the recommended difficulty.

\subsection{Conclusion}

World-model rollouts support intervention decisions only if they change difficulty for the same target population.
On our paired-intervention benchmark, CaLA improves population estimates over unadjusted rollouts under DDA confounding and reduces individual regret under both assignment rules.
Randomized and oracle controls reveal limits in estimating churn from logs or residual-corrected rollouts.
Before using rollouts to choose difficulty, we must test their churn predictions when setting difficulty for the same players, not only their next-state accuracy.

\section*{Ethics Statement}
All datasets and player histories used in this study are generated by an explicit simulator. The work uses no human participants or personal data. The simulator is a controlled scientific instrument and should not be interpreted as evidence about the behavior of real players or the effects of deployed commercial game systems. A churn-minimizing difficulty policy could nevertheless affect player autonomy, enjoyment, and fairness if deployed. Any deployment should evaluate heterogeneous effects, constrain optimization objectives beyond retention, avoid manipulative difficulty changes, and validate transport from the synthetic structural model to the intended population.

\section*{Reproducibility Statement}
The supplementary material documents the simulator, dataset construction, player-disjoint splits, model architectures, training protocol, evaluation metrics, and statistical units. The synthetic data can be regenerated with the provided engine and simulator code, and all reported comparisons use fixed splits and three training seeds. The anonymized codebase, dependency specification, and reproduction instructions will be provided at \url{https://anonymous.4open.science/r/Match3-CaLA/} and included in the supplementary materials under the Apache License 2.0. All model training and evaluation runs used a single NVIDIA A100 GPU with 40 GB of memory.

\section*{AI use statement}
Generative AI tools were used to assist with software implementation, debugging, code review, and \LaTeX{} formatting. AI tools were also used to edit the manuscript for language, wording, and phrasing to improve readability. The authors reviewed all code, analyses, and text and take full responsibility for the final content of this work.

\bibliography{references}

@article{reguera2020quantifying,
  title={Quantifying human engagement into playful activities},
  author={Reguera, David and Colomer-de-Sim{\'o}n, Pol and Encinas, Iv{\'a}n and Sort, Manel and Wedekind, Jan and Bogu{\~n}{\'a}, Mari{\'a}n},
  journal={Scientific Reports},
  volume={10},
  number={1},
  pages={4145},
  year={2020},
  publisher={Nature Publishing Group UK London}
}

@inproceedings{kristensen2021,
  title     = {Statistical Modelling of Level Difficulty in Puzzle Games},
  author    = {Kristensen, Jeppe Theiss and Valdivia, Arturo and Burelli, Paolo},
  booktitle = {2021 IEEE Conference on Games (CoG)},
  pages     = {1--8},
  publisher = {IEEE},
  year      = {2021},
  doi       = {10.1109/CoG52621.2021.9619050}
}

@article{roohi2021,
  title   = {Predicting Game Difficulty and Engagement Using AI Players},
  author  = {Roohi, Shaghayegh and Guckelsberger, Christian and Relas, Asko and
             Heiskanen, Henri and Takatalo, Jari and H{\"a}m{\"a}l{\"a}inen, Perttu},
  journal = {Proceedings of the ACM on Human-Computer Interaction},
  volume  = {5},
  number  = {CHI PLAY},
  pages   = {1--17},
  publisher = {Association for Computing Machinery},
  year    = {2021},
  doi     = {10.1145/3474658}
}

@inproceedings{xue2017,
  title     = {Dynamic Difficulty Adjustment for Maximized Engagement in Digital Games},
  author    = {Xue, Su and Wu, Meng and Kolen, John and Aghdaie, Navid and
               Zaman, Kazi A.},
  booktitle = {Proceedings of the 26th International Conference on World Wide Web
               Companion},
  pages     = {465--471},
  publisher = {ACM},
  year      = {2017},
  doi       = {10.1145/3041021.3054170}
}

@article{cutting2023,
  title   = {Difficulty-Skill Balance Does Not Affect Engagement and Enjoyment:
             A Pre-Registered Study Using Artificial Intelligence-Controlled
             Difficulty},
  author  = {Cutting, Joe and Deterding, Sebastian and Demediuk, Simon and
             Sephton, Nick},
  journal = {Royal Society Open Science},
  volume  = {10},
  number  = {2},
  pages   = {220274},
  year    = {2023},
  doi     = {10.1098/rsos.220274}
}

@inproceedings{louizos2017,
  title     = {Causal Effect Inference with Deep Latent-Variable Models},
  author    = {Louizos, Christos and Shalit, Uri and Mooij, Joris and
               Sontag, David and Zemel, Richard and Welling, Max},
  booktitle = {Advances in Neural Information Processing Systems},
  volume    = {30},
  publisher = {Curran Associates, Inc.},
  year      = {2017},
  url       = {https://proceedings.neurips.cc/paper_files/paper/2017/file/94b5bde6de888ddf9cde6748ad2523d1-Paper.pdf}
}

@inproceedings{assran2023ijepa,
  title     = {Self-Supervised Learning from Images with a Joint-Embedding
               Predictive Architecture},
  author    = {Assran, Mahmoud and Duval, Quentin and Misra, Ishan and
               Bojanowski, Piotr and Vincent, Pascal and Rabbat, Michael and
               LeCun, Yann and Ballas, Nicolas},
  booktitle = {Proceedings of the IEEE/CVF Conference on Computer Vision and
               Pattern Recognition},
  pages     = {15619--15629},
  month     = jun,
  year      = {2023},
  doi       = {10.1109/CVPR52729.2023.01499},
  url       = {https://openaccess.thecvf.com/content/CVPR2023/html/Assran_Self-Supervised_Learning_From_Images_With_a_Joint-Embedding_Predictive_Architecture_CVPR_2023_paper.html}
}

@article{balestriero2025lejepa,
  title         = {{LeJEPA}: Provable and Scalable Self-Supervised Learning
                   Without the Heuristics},
  author        = {Balestriero, Randall and LeCun, Yann},
  journal       = {arXiv preprint arXiv:2511.08544},
  year          = {2025},
  eprint        = {2511.08544},
  archivePrefix = {arXiv},
  primaryClass  = {cs.LG},
  url           = {https://arxiv.org/abs/2511.08544}
}

@inproceedings{kingma2014vae,
  title     = {Auto-Encoding Variational Bayes},
  author    = {Kingma, Diederik P. and Welling, Max},
  booktitle = {2nd International Conference on Learning Representations
               (ICLR)},
  year      = {2014},
  eprint    = {1312.6114},
  archivePrefix = {arXiv},
  primaryClass  = {stat.ML},
  url       = {https://arxiv.org/abs/1312.6114}
}

@inproceedings{vaswani2017attention,
  title     = {Attention Is All You Need},
  author    = {Vaswani, Ashish and Shazeer, Noam and Parmar, Niki and
               Uszkoreit, Jakob and Jones, Llion and Gomez, Aidan N. and
               Kaiser, {\L}ukasz and Polosukhin, Illia},
  booktitle = {Advances in Neural Information Processing Systems},
  volume    = {30},
  publisher = {Curran Associates, Inc.},
  year      = {2017},
  url       = {https://proceedings.neurips.cc/paper_files/paper/2017/file/3f5ee243547dee91fbd053c1c4a845aa-Paper.pdf}
}

@inproceedings{cho2014gru,
  title     = {Learning Phrase Representations Using {RNN}
               Encoder--Decoder for Statistical Machine Translation},
  author    = {Cho, Kyunghyun and van Merri{\"e}nboer, Bart and
               Gulcehre, Caglar and Bahdanau, Dzmitry and Bougares, Fethi and
               Schwenk, Holger and Bengio, Yoshua},
  editor    = {Moschitti, Alessandro and Pang, Bo and Daelemans, Walter},
  booktitle = {Proceedings of the 2014 Conference on Empirical Methods in
               Natural Language Processing (EMNLP)},
  address   = {Doha, Qatar},
  publisher = {Association for Computational Linguistics},
  pages     = {1724--1734},
  month     = oct,
  year      = {2014},
  doi       = {10.3115/v1/D14-1179},
  url       = {https://aclanthology.org/D14-1179/}
}

@inproceedings{loshchilov2019adamw,
  title     = {Decoupled Weight Decay Regularization},
  author    = {Loshchilov, Ilya and Hutter, Frank},
  booktitle = {International Conference on Learning Representations},
  year      = {2019},
  url       = {https://openreview.net/forum?id=Bkg6RiCqY7}
}

@article{gneiting2007scoring,
  title     = {Strictly Proper Scoring Rules, Prediction, and Estimation},
  author    = {Gneiting, Tilmann and Raftery, Adrian E.},
  journal   = {Journal of the American Statistical Association},
  volume    = {102},
  number    = {477},
  pages     = {359--378},
  publisher = {Informa UK Limited},
  year      = {2007},
  doi       = {10.1198/016214506000001437}
}

@book{pearl2009,
  title     = {Causality: Models, Reasoning, and Inference},
  author    = {Pearl, Judea},
  edition   = {2nd},
  publisher = {Cambridge University Press},
  year      = {2009}
}

@incollection{hirano2004,
  title     = {The Propensity Score with Continuous Treatments},
  author    = {Hirano, Keisuke and Imbens, Guido W.},
  booktitle = {Applied Bayesian Modeling and Causal Inference from
               Incomplete-Data Perspectives},
  editor    = {Gelman, Andrew and Meng, Xiao-Li},
  pages     = {73--84},
  publisher = {Wiley},
  year      = {2004},
  doi       = {10.1002/0470090456.ch7}
}

@article{kennedy2017,
  title   = {Non-parametric Methods for Doubly Robust Estimation of Continuous
             Treatment Effects},
  author  = {Kennedy, Edward H. and Ma, Zongming and McHugh, Matthew D. and
             Small, Dylan S.},
  journal = {Journal of the Royal Statistical Society: Series B (Statistical
             Methodology)},
  volume  = {79},
  number  = {4},
  pages   = {1229--1245},
  year    = {2017},
  doi     = {10.1111/rssb.12212}
}

@article{chernozhukov2018,
  title   = {Double/Debiased Machine Learning for Treatment and Structural
             Parameters},
  author  = {Chernozhukov, Victor and Chetverikov, Denis and Demirer, Mert and
             Duflo, Esther and Hansen, Christian and Newey, Whitney and
             Robins, James},
  journal = {The Econometrics Journal},
  volume  = {21},
  number  = {1},
  pages   = {C1--C68},
  year    = {2018},
  doi     = {10.1111/ectj.12097}
}

@inproceedings{ha2018,
  title     = {Recurrent World Models Facilitate Policy Evolution},
  author    = {Ha, David and Schmidhuber, J{\"u}rgen},
  booktitle = {Advances in Neural Information Processing Systems},
  volume    = {31},
  publisher = {Curran Associates, Inc.},
  year      = {2018}
}

@article{rezende2020,
  title   = {Causally Correct Partial Models for Reinforcement Learning},
  author  = {Rezende, Danilo J. and Danihelka, Ivo and Papamakarios, George and Ke, Nan Rosemary and Jiang, Ray and Weber, Th{\'e}ophane and Gregor, Karol and Merzic, Hamza and Viola, Fabio and Wang, Jane and Mitrovic, Jovana and Besse, Frederic and Antonoglou, Ioannis and Buesing, Lars},
  journal = {arXiv preprint arXiv:2002.02836},
  year    = {2020}
}

@incollection{bareinboim2022,
  title     = {On {Pearl}'s Hierarchy and the Foundations of Causal Inference},
  author    = {Bareinboim, Elias and Correa, Juan D. and Ibeling, Duligur and Icard, Thomas},
  booktitle = {Probabilistic and Causal Inference: The Works of Judea Pearl},
  pages     = {507--556},
  publisher = {ACM},
  year      = {2022},
  doi       = {10.1145/3501714.3501743}
}

@inproceedings{buesing2019,
  title     = {Woulda, Coulda, Shoulda: Counterfactually-Guided Policy Search},
  author    = {Buesing, Lars and Weber, Th{\'e}ophane and Zwols, Yori and Racani{\`e}re, S{\'e}bastien and Guez, Arthur and Lespiau, Jean-Baptiste and Heess, Nicolas},
  booktitle = {International Conference on Learning Representations},
  year      = {2019},
  url       = {https://openreview.net/forum?id=BJG0voC9YQ}
}

@inproceedings{bica2020deconfounder,
  title     = {Time Series Deconfounder: Estimating Treatment Effects over Time in the Presence of Hidden Confounders},
  author    = {Bica, Ioana and Alaa, Ahmed M. and van der Schaar, Mihaela},
  booktitle = {Proceedings of the 37th International Conference on Machine Learning},
  pages     = {884--895},
  volume    = {119},
  series    = {Proceedings of Machine Learning Research},
  publisher = {PMLR},
  year      = {2020}
}

@inproceedings{bica2020crn,
  title     = {Estimating Counterfactual Treatment Outcomes over Time through Adversarially Balanced Representations},
  author    = {Bica, Ioana and Alaa, Ahmed M. and Jordon, James and van der Schaar, Mihaela},
  booktitle = {International Conference on Learning Representations},
  year      = {2020},
  url       = {https://openreview.net/forum?id=BJg866NFvB}
}

@inproceedings{melnychuk2022,
  title     = {Causal Transformer for Estimating Counterfactual Outcomes},
  author    = {Melnychuk, Valentyn and Frauen, Dennis and Feuerriegel, Stefan},
  booktitle = {Proceedings of the 39th International Conference on Machine Learning},
  pages     = {15293--15329},
  volume    = {162},
  series    = {Proceedings of Machine Learning Research},
  publisher = {PMLR},
  year      = {2022}
}

@inproceedings{damour2019,
  title     = {On Multi-Cause Causal Inference with Unobserved Confounding: Counterexamples, Impossibility, and Alternatives},
  author    = {D'Amour, Alexander},
  booktitle = {Proceedings of the Twenty-Second International Conference on Artificial Intelligence and Statistics},
  pages     = {3478--3486},
  volume    = {89},
  series    = {Proceedings of Machine Learning Research},
  publisher = {PMLR},
  year      = {2019}
}

@article{miao2018,
  title   = {Identifying Causal Effects with Proxy Variables of an Unmeasured Confounder},
  author  = {Miao, Wang and Geng, Zhi and Tchetgen Tchetgen, Eric J.},
  journal = {Biometrika},
  volume  = {105},
  number  = {4},
  pages   = {987--993},
  year    = {2018},
  doi     = {10.1093/biomet/asy038}
}

@inproceedings{rissanen2021,
  title     = {A Critical Look at the Consistency of Causal Estimation with Deep Latent Variable Models},
  author    = {Rissanen, Severi and Marttinen, Pekka},
  booktitle = {Advances in Neural Information Processing Systems},
  pages     = {4207--4217},
  volume    = {34},
  publisher = {Curran Associates, Inc.},
  year      = {2021}
}

@inproceedings{shalit2017,
  title     = {Estimating Individual Treatment Effect: Generalization Bounds and Algorithms},
  author    = {Shalit, Uri and Johansson, Fredrik D. and Sontag, David},
  booktitle = {Proceedings of the 34th International Conference on Machine Learning},
  pages     = {3076--3085},
  volume    = {70},
  series    = {Proceedings of Machine Learning Research},
  publisher = {PMLR},
  year      = {2017}
}

@inproceedings{jiang2016,
  title     = {Doubly Robust Off-Policy Value Evaluation for Reinforcement Learning},
  author    = {Jiang, Nan and Li, Lihong},
  booktitle = {Proceedings of The 33rd International Conference on Machine Learning},
  pages     = {652--661},
  volume    = {48},
  series    = {Proceedings of Machine Learning Research},
  publisher = {PMLR},
  year      = {2016}
}

\section*{Checklist}

\begin{enumerate}
    \item For all models and algorithms presented, check if you include:
    \begin{enumerate}
        \item A clear description of the mathematical setting, assumptions, algorithm, and/or model. \textbf{Yes.} Sections~3--4 and the supplementary material specify the SCM, factorization, estimators, and rollout procedure.
        \item An analysis of the properties and complexity (time, space, sample size) of any algorithm. \textbf{No.} The paper reports sample sizes and evaluation budgets but does not provide a formal time or space complexity analysis.
        \item (Optional) Anonymized source code, with specification of all dependencies, including external libraries. \textbf{Yes.} The anonymized codebase and dependency specification will be provided at \url{https://anonymous.4open.science/r/Match3-CaLA/} and included in the supplementary materials.
    \end{enumerate}

    \item For any theoretical claim, check if you include:
    \begin{enumerate}
        \item Statements of the full set of assumptions of all theoretical results. \textbf{Not Applicable.} The paper presents an empirical method and does not state formal theoretical results.
        \item Complete proofs of all theoretical results. \textbf{Not Applicable.}
        \item Clear explanations of any assumptions. \textbf{Yes.} The identifying and structural assumptions are stated in Section~4.
    \end{enumerate}

    \item For all figures and tables that present empirical results, check if you include:
    \begin{enumerate}
        \item The code, data, and instructions needed to reproduce the main experimental results (either in the supplemental material or as a URL). \textbf{Yes.} The supplementary materials and anonymized repository at \url{https://anonymous.4open.science/r/Match3-CaLA/} provide the code, synthetic data generator, dependencies, and reproduction instructions.
        \item All the training details (e.g., data splits, hyperparameters, how they were chosen). \textbf{Yes.} These details are provided in Section~5 and the supplementary material.
        \item A clear definition of the specific measure or statistics and error bars (e.g., with respect to the random seed after running experiments multiple times). \textbf{Yes.} Metrics and aggregation across three training seeds are defined in Section~5 and the supplementary material.
        \item A description of the computing infrastructure used. (e.g., type of GPUs, internal cluster, or cloud provider). \textbf{Yes.} All model training and evaluation runs used a single NVIDIA A100 GPU with 40 GB of memory.
    \end{enumerate}

    \item If you are using existing assets (e.g., code, data, models) or curating/releasing new assets, check if you include:
    \begin{enumerate}
        \item Citations of the creator if your work uses existing assets. \textbf{Yes.} Existing methods and model families are cited where introduced.
        \item The license information of the assets, if applicable. \textbf{Yes.} The released code and synthetic-data assets are provided under the Apache License 2.0.
        \item New assets either in the supplemental material or as a URL, if applicable. \textbf{Yes.} The synthetic dataset and its generator are described as supplementary assets.
        \item Information about consent from data providers/curators. \textbf{Not Applicable.} All data are generated by the simulator and contain no human-player data.
        \item Discussion of sensitive content if applicable, e.g., personally identifiable information or offensive content. \textbf{Yes.} The Ethics Statement explains that the study uses no personal data and discusses deployment risks.
    \end{enumerate}

    \item If you used crowdsourcing or conducted research with human subjects, check if you include:
    \begin{enumerate}
        \item The full text of instructions given to participants and screenshots. \textbf{Not Applicable.}
        \item Descriptions of potential participant risks, with links to Institutional Review Board (IRB) approvals if applicable. \textbf{Not Applicable.}
        \item The estimated hourly wage paid to participants and the total amount spent on participant compensation. \textbf{Not Applicable.}
    \end{enumerate}
\end{enumerate}

\clearpage
\onecolumn
\appendix
\aistatstitle{Supplementary Materials}
\suppressfloats[t]

\section{Data, Simulation, and Evaluation Protocol}
\label{app:data-details}

\subsection{Dataset and Structure}

Across all experiments, we used the synthetic Match-3 dataset ``wrong-move-reference,'' which is made available in the supplementary materials and can also be generated using the provided engine and simulator code. All experiments used a fixed release and player partition across training seeds. The dataset contains simulated players. No human-player data were used.

The overall dataset contains longitudinal histories for the same catalog of 2,400 player identities under natural and randomized difficulty assignment. Each regime is stored in ten shards of 240 players, made available in the supplementary materials and to be released publicly after the review phase. The two regimes are separate realizations of the same simulated population and should not be interpreted as 4,800 independent player identities. Their realized actions, outcomes, and survival to later attempts may differ.

The data are organized at three nested levels:
\begin{enumerate}[label=\arabic*)]
    \item \textbf{Player level:} a stable player identifier links an ordered history of attempts and the player's status.

    \item \textbf{Attempt level:} each record contains the level family, baseline tier, served difficulty, completion, completion margin, churn, and post-attempt behavioral evidence.

    \item \textbf{Transition level:} each record contains the current color and special-kind grids, counters, selected action, and resulting settled state. A transition includes the complete resolution of the swap, including cascades, gravity, and refill.
\end{enumerate}

Attempts are typically contiguous within a player history and stop at churn or administrative censoring after attempt $20$. Transition indices restart within each attempt. Table~\ref{tab:release-overview} gives the overall counts and related quantities.

\begin{table}[t]
\caption{Complete-release counts. Active players have an observed attempt-20 opening.}
\label{tab:release-overview}
\centering
\scriptsize
\begin{tabular}{lrrrr}
\toprule
    \textbf{Regime} & \textbf{Players} & \textbf{Attempts} & \textbf{Transitions} & \shortstack{\textbf{Active at}\\\textbf{attempt 20}} \\
    \midrule
    Natural
    & 2,400 & 47,810 & 878,197 & 2,377 \\
    Randomized
    & 2,400 & 47,645 & 841,434 & 2,362 \\
\bottomrule
\end{tabular}
\end{table}

\subsection{Level Families and Assignment Regimes}

As highlighted in the main paper, all three level families use an $8\times8$ board and the same underlying match, cascade, gravity, and refill mechanics. Orchard uses five colors with uniform refill, Harbour uses five colors with nonuniform refill, and Foundry uses six colors with uniform refill. The special-kind grid distinguishes ordinary tiles from horizontal and vertical striped tiles.

Baseline tier and served difficulty are distinct variables. The easy, medium, and hard tiers determine move budgets of $22$, $20$, and $18$, respectively. Served difficulty changes the goal quota through the calibrated difficulty-to-quota map. Under natural assignment, served difficulty depends on the player's simulator skill and the level-specific skill demands. ``Stronger players'' therefore tend to receive harder content. Under randomized assignment, the skill-dependent assignment coefficients are set to zero and assignment noise is adjusted to approximately preserve the marginal difficulty distribution within level and tier.

The resulting comparison changes the assignment mechanism while retaining the benchmark's player catalog and gameplay mechanisms. Randomization does not remove player heterogeneity in behavior or response.

\subsection{Player-Disjoint Splits and Landmark Populations}

Players are partitioned into three sets containing $1,680$ training, $360$ validation, and $360$ test identities per regime, corresponding to a 70/15/15 split. The partition is defined before history-prefix examples or normalization statistics are constructed. All attempts and transitions belonging to a player remain in the same split. For all experiments, we use three seeds and report the mean and corresponding sample SD.

The causal evaluation takes place immediately before the breakpoint of attempt $20$. A player is eligible if they remain active and have a logged opening for that attempt. Consequently, the held-out test evaluation contains $357$ natural-assignment players and $355$ randomized-assignment players, rather than all $360$ nominal test identities. Table~\ref{tab:player-split-accounting} distinguishes the nominal splits from the active landmark cohorts.

\begin{table}[t]
\caption{Player counts by split and eligibility at attempt $20$.}
\label{tab:player-split-accounting}
\centering
\scriptsize
\begin{tabular}{llrr}
\toprule
\textbf{Regime} & \textbf{Split} & \textbf{Nominal players} & \shortstack{\textbf{Active at}\\\textbf{attempt 20}} \\
\midrule
Natural & Training   & 1,680 & 1,660 \\
Natural & Validation & 360   & 360 \\
Natural & Test       & 360   & 357 \\
\midrule
Randomized & Training   & 1,680 & 1,650 \\
Randomized & Validation & 360   & 357 \\
Randomized & Test       & 360   & 355 \\
\bottomrule
\end{tabular}
\end{table}

Each eligible player contributes only to the level logged at the landmark. Players are not replicated across all level families. Table~\ref{tab:landmark-composition} reports the resulting composition and observed test churn. These churn frequencies describe outcomes under the logging regime and are not estimates of risk under a fixed difficulty intervention.

\begin{table}[t]
\caption{Landmark cohort composition and observed test churn after attempt $20$.}
\label{tab:landmark-composition}
\centering
\scriptsize
\begin{tabular}{llrrrr}
\toprule
\textbf{Regime} & \textbf{Level} & \textbf{Validation $n$} & \textbf{Test $n$}
& \textbf{Churn count} & \textbf{Churn (\%)} \\
\midrule
Natural & Orchard & 160 & 181 & 12 & 6.63 \\
Natural & Harbour & 122 & 105 & 17 & 16.19 \\
Natural & Foundry & 78  & 71  & 13 & 18.31 \\
Natural & Total   & 360 & 357 & 42 & 11.76 \\
\midrule
Randomized & Orchard & 159 & 181 & 25 & 13.81 \\
Randomized & Harbour & 122 & 104 & 31 & 29.81 \\
Randomized & Foundry & 76  & 70  & 26 & 37.14 \\
Randomized & Total   & 357 & 355 & 82 & 23.10 \\
\bottomrule
\end{tabular}
\end{table}

\subsection{World-Model Data and Predictive Evaluation}

As explained in Appendix~\ref{app:world-model-architecture-and-rollout-evidence}, the world-model dynamics are fitted using natural-assignment training transitions and selected using natural-assignment validation data. The selected dynamics checkpoints are then frozen and reused for both assignment regimes. An episode in the world-model pipeline denotes one attempt, not an entire player history. Table~\ref{tab:wm-data-accounting} itemizes the data used for world-model fitting and validation. Mask-valid transitions exclude attempts containing a reshuffle event and transitions with inconsistent settled-pair reconstructions, enumerated in the two bottom rows.

\begin{table}[t]
\caption{Natural-assignment data used for world-modeling.}
\label{tab:wm-data-accounting}
\centering
\scriptsize
\begin{tabular}{lrr}
\toprule
\textbf{Quantity} & \textbf{Training} & \textbf{Validation} \\
\midrule
Players & 1,680 & 360 \\
Attempt episodes & 33,448 & 7,200 \\
Transitions & 614,829 & 131,929 \\
Mean transitions per episode & 18.38 & 18.32 \\
Maximum transitions per episode & 22 & 22 \\
Mask-valid transitions & 613,852 & 131,631 \\
Transitions flagged as stripe-present & 191,264 & 41,561 \\
Transitions with inferred stripe activation & 26,709 & 5,557 \\
Transitions in attempts with reshuffling & 924 & 289 \\
Inconsistent settled-pair reconstructions & 53 & 9 \\
\bottomrule
\end{tabular}
\end{table}

The architecture comparison of the considered world-model families uses a fixed subset of $2,000$ validation episodes comprising 36,659 transitions. Logged-action replay evaluates horizons $k\in\{1,3,7\}$. Engine-conditioned and autonomous-imagination diagnostics use up to $512$ validation anchors and four rollouts per anchor. Additional details about the modeling and evaluation procedures are provided in Appendix~\ref{app:world-model-architecture-and-rollout-evidence}.

\subsection{Temporal Availability and Preprocessing}

For a target attempt $j$, the inferred player representation uses only completed attempts strictly before attempt $j$. At the evaluation landmark, this history is $H_{i,<20}$, which consists of attempts $1$ through $19$. Target difficulty, target gameplay, completion, margin, post-attempt evidence, and churn do not enter the history encoder, thereby preventing leakage.

The historical inputs include color boards, actions, counters, level and tier, previous difficulties, completion outcomes, margins, and 12 other synthetic behavioral indicators: search latency, candidate recall, hint count, pattern-error rate, immediate-pattern precision, distractor resistance, lookahead-choice rate, setup value, cascade preparation, goal-clear share, goals per move, and moves-left efficiency. Post-attempt indicators become eligible only for subsequent decisions. Consequently, the representation uses gameplay and simulator-generated behavioral proxies rather than board trajectories alone.

The observable base context $b_i\in\mathbb{R}^{7}$ combines level and tier one-hot vectors with experience reconstructed from previous completion outcomes. Handcrafted adjustment adds 40 history summaries, standardized using training data. Learned adjustment adds four latent coordinates.

The target opening is used to initialize the intervention rollout but is not included in posterior inference from the historical prefix. The target outcomes and post-opening transitions remain unavailable during prediction generation. The logged test outcomes are accessed after predictions are frozen for evaluation and the population WM-DR correction. They are not inputs to individualized recommendations.

\subsection{Engine Reference and Statistical Units}

The considered intervention grid for our experiments contains $17$ candidate difficulties: $\mathcal{E}=\{-2,-1.75,\ldots,1.75,2\}.$ The corrected engine reference uses each test player's logged attempt-20 opening, level, tier, reconstructed experience, and oracle skill. It propagates striped-special state and uses $16$ paired engine replicates per player. The proposed variational CaLA uses 16 posterior draws and eight trajectory replicates per draw. Nonvariational adjustment contexts use one effective context draw. Posterior samples, rollout replicates, and candidate difficulties are repeated computations for the same player, not additional independent observations.

The doubly robust estimators use a Gaussian kernel with bandwidth $h=0.75$, a propensity density floor $g_{\min}=0.02$, and an individual-support threshold of $0.01$ on the fitted propensity density. All three values were fixed before test evaluation.

Results are reported as the mean and corresponding sample SD across the three training seeds. Population curve MAE is weighted by level-specific player counts. Population regret averages the three level-specific decisions equally. Individualized regret pools players within each seed. The causal evaluation uses held-out player identities from the benchmark.

\subsection{Structural Causal Model Graphs}\label{app:scm-graphs}

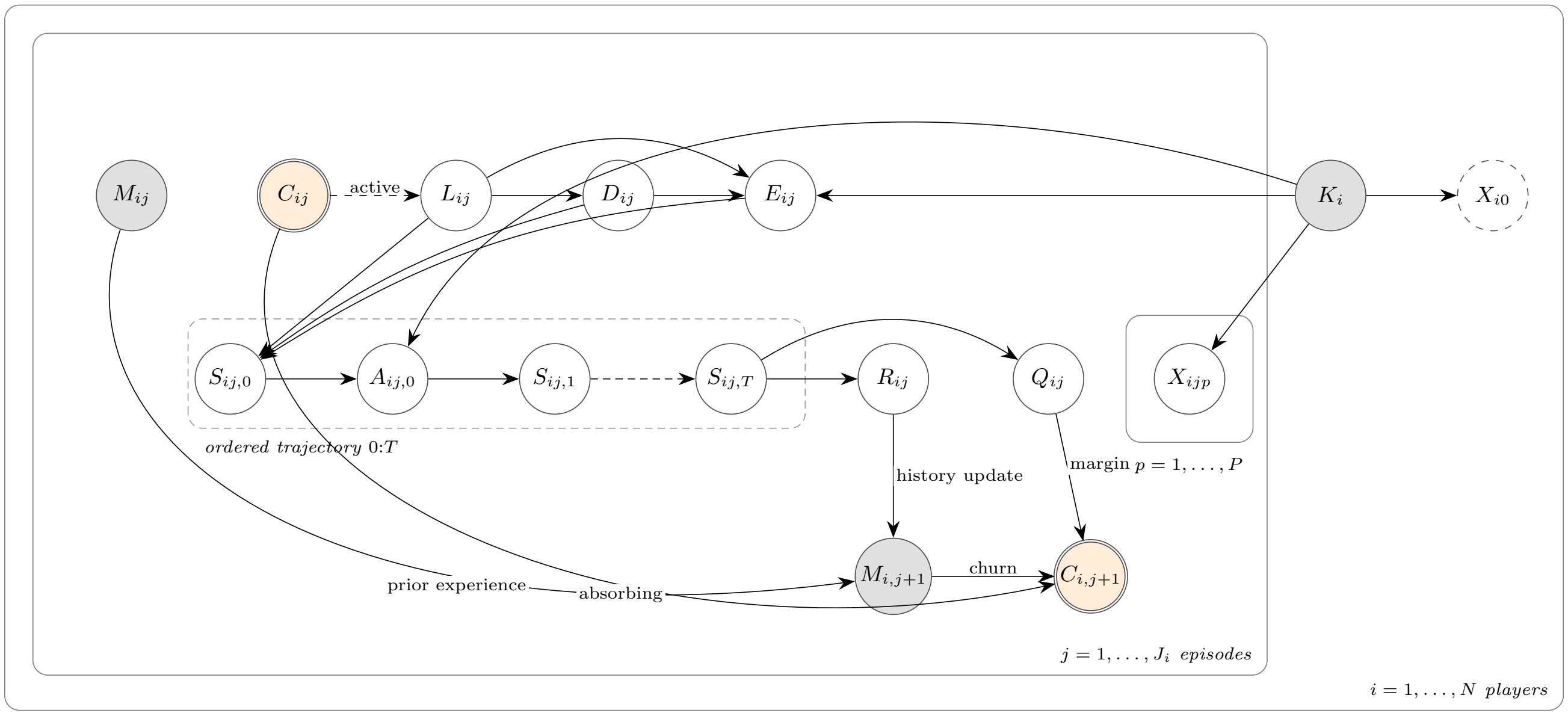
\begin{figure}[t]
\centering
\resizebox{\textwidth}{!}{%
\begin{tikzpicture}[
  >={Stealth[length=2.4mm]},
  var/.style={circle, draw=black!70, fill=white, minimum size=10mm, inner sep=1pt, font=\small},
  det/.style={var, fill=black!12},
  churn/.style={var, double, double distance=0.6pt, fill=orange!15},
  optvar/.style={var, dashed},
  lbl/.style={font=\scriptsize, fill=white, inner sep=1pt},
  plate/.style={draw=black!50, rounded corners=6pt},
  platelbl/.style={font=\scriptsize\itshape}
]
\node[det]    (M)  at (0,4.6)   {$M_{ij}$};
\node[churn]  (C)  at (2.3,4.6) {$C_{ij}$};
\node[var]    (L)  at (4.6,4.6) {$L_{ij}$};
\node[var]    (D)  at (6.9,4.6) {$D_{ij}$};
\node[var]    (E)  at (9.2,4.6) {$E_{ij}$};
\node[det]    (K)  at (17.0,4.6){$K_i$};
\node[optvar] (X0) at (19.3,4.6){$X_{i0}$};
\node[var] (S0) at (1.4,2.0)  {$S_{ij,0}$};
\node[var] (A0) at (3.7,2.0)  {$A_{ij,0}$};
\node[var] (S1) at (6.0,2.0)  {$S_{ij,1}$};
\node[var] (ST) at (8.5,2.0)  {$S_{ij,T}$};
\node[var] (R)  at (10.8,2.0) {$R_{ij}$};
\node[var] (Q)  at (13.0,2.0) {$Q_{ij}$};
\node[var] (Xp) at (15.0,2.0) {$X_{ijp}$};
\node[det]   (Mn) at (10.8,-0.8) {$M_{i,j+1}$};
\node[churn] (Cn) at (13.6,-0.8) {$C_{i,j+1}$};
\draw[plate, densely dashed, draw=black!40] (0.8,1.3) rectangle (9.55,2.85);
\node[platelbl, anchor=north west] at (0.9,1.25) {ordered trajectory $0{:}T$};
\draw[plate] (14.1,1.1) rectangle (15.9,2.9);
\node[platelbl, anchor=north west] at (14.1,1.0) {$p=1,\dots,P$};
\draw[plate] (-1.4,-2.2) rectangle (16.1,6.9);
\node[platelbl, anchor=south east] at (16.0,-2.15) {$j=1,\dots,J_i$ episodes};
\draw[plate] (-1.8,-2.7) rectangle (20.3,7.3);
\node[platelbl, anchor=south east] at (20.2,-2.65) {$i=1,\dots,N$ players};
\draw[->, dashed] (C) -- node[lbl, above]{active} (L);
\draw[->] (L) -- (D);
\draw[->] (D) -- (E);
\draw[->] (L) to[bend left=30] (E);
\draw[->] (L) -- (S0);
\draw[->] (D) to[bend right=10] (S0);
\draw[->] (E) to[bend right=14] (S0);
\draw[->] (S0) -- (A0);
\draw[->] (A0) -- (S1);
\draw[->, densely dashed] (S1) -- (ST);
\draw[->] (ST) -- (R);
\draw[->] (ST) to[bend left=32] (Q);
\draw[->] (K) -- (E);
\draw[->] (K) -- (X0);
\draw[->] (K) -- (Xp);
\draw[->] (K) .. controls (11.5,6.4) and (5.5,5.8) .. (A0);
\draw[->] (R) -- node[lbl, right]{history update} (Mn);
\draw[->] (Q) -- node[lbl, right, pos=0.4]{margin} (Cn);
\draw[->] (Mn) -- node[lbl, above]{churn} (Cn);
\draw[->] (M) .. controls (-1.2,1.0) and (3.0,-1.8) .. node[lbl, below, pos=0.7]{prior experience} (Mn);
\draw[->] (C) .. controls (0.6,0.6) and (7.0,-2.2) .. node[lbl, below, pos=0.65]{absorbing} (Cn);
\end{tikzpicture}%
%
}
\caption{The Match-3 SCM across player, attempt, and gameplay scales. Oracle player state $K_i$ is shared across attempts. Level $L_{ij}$, tier $D_{ij}$, and served difficulty $E_{ij}$ initialize the state-action trajectory, and together with $K_i$ they condition every transition and action along it, with only the first such arrows drawn. Completion $R_{ij}$ updates expected experience from $M_{ij}$ to $M_{i,j+1}$, and the updated history and signed margin $Q_{ij}$ drive next-attempt churn $C_{i,j+1}$. Gray nodes are deterministic or oracle-only quantities, double circles mark the absorbing churn indicators, the dashed node $X_{i0}$ is the optional baseline assessment, and dashed arrows mark elided Markov steps and the activity gate.}
\label{fig:scm-plate-dag}
\end{figure}

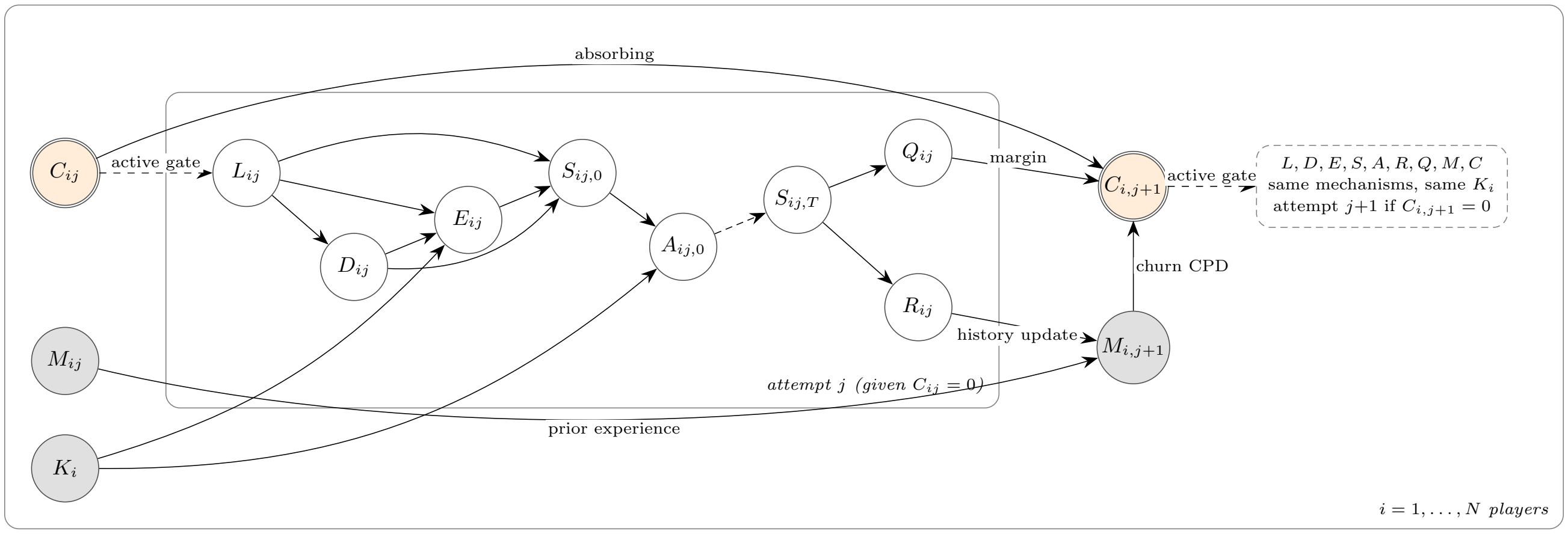
\begin{figure}[t]
\centering
\resizebox{\textwidth}{!}{%
\begin{tikzpicture}[
  >={Stealth[length=2.4mm]},
  var/.style={circle, draw=black!70, fill=white, minimum size=10mm, inner sep=1pt, font=\small},
  det/.style={var, fill=black!12},
  churn/.style={var, double, double distance=0.6pt, fill=orange!15},
  lbl/.style={font=\scriptsize, fill=white, inner sep=1pt},
  plate/.style={draw=black!50, rounded corners=6pt},
  platelbl/.style={font=\scriptsize\itshape}
]
\node[churn] (C) at (0,4.4) {$C_{ij}$};
\node[det]   (M) at (0,1.6) {$M_{ij}$};
\node[det]   (K) at (0,0.0) {$K_i$};
\draw[plate] (1.5,0.9) rectangle (13.9,5.6);
\node[platelbl, anchor=south east] at (13.8,0.98) {attempt $j$ (given $C_{ij}=0$)};
\node[var] (L)  at (2.7,4.4)  {$L_{ij}$};
\node[var] (D)  at (4.3,3.0)  {$D_{ij}$};
\node[var] (E)  at (6.0,3.7)  {$E_{ij}$};
\node[var] (S0) at (7.7,4.4)  {$S_{ij,0}$};
\node[var] (A0) at (9.2,3.3)  {$A_{ij,0}$};
\node[var] (ST) at (10.9,4.0) {$S_{ij,T}$};
\node[var] (Q)  at (12.7,4.7) {$Q_{ij}$};
\node[var] (R)  at (12.7,2.4) {$R_{ij}$};
\node[det]   (Mn) at (15.9,1.8) {$M_{i,j+1}$};
\node[churn] (Cn) at (15.9,4.2) {$C_{i,j+1}$};
\node[plate, densely dashed, align=center, font=\scriptsize, inner sep=5pt] (next) at (19.6,4.2) {$L,D,E,S,A,R,Q,M,C$\\[1pt] same mechanisms, same $K_i$\\[1pt] attempt $j{+}1$ if $C_{i,j+1}=0$};
\draw[plate] (-0.9,-0.9) rectangle (22.3,6.9);
\node[platelbl, anchor=south east] at (22.2,-0.85) {$i=1,\dots,N$ players};
\draw[->, dashed] (C) -- node[lbl, above]{active gate} (L);
\draw[->] (C) .. controls (4.5,6.5) and (12,6.5) .. node[lbl, above, pos=0.5]{absorbing} (Cn);
\draw[->] (L) -- (D);
\draw[->] (L) -- (E);
\draw[->] (D) -- (E);
\draw[->] (L) to[bend left=20] (S0);
\draw[->] (E) -- (S0);
\draw[->] (D) to[bend right=25] (S0);
\draw[->] (S0) -- (A0);
\draw[->, densely dashed] (A0) -- (ST);
\draw[->] (ST) -- (Q);
\draw[->] (ST) -- (R);
\draw[->] (K) to[bend right=15] (E);
\draw[->] (K) to[bend right=20] (A0);
\draw[->] (M) .. controls (5.0,0.4) and (11.5,0.5) .. node[lbl, below, pos=0.5]{prior experience} (Mn);
\draw[->] (Q) -- node[lbl, above, pos=0.45]{margin} (Cn);
\draw[->] (R) -- node[lbl, below, pos=0.45]{history update} (Mn);
\draw[->] (Mn) -- node[lbl, right]{churn CPD} (Cn);
\draw[->, dashed] (Cn) -- node[lbl, above]{active gate} (next);
\end{tikzpicture}%
%
}
\caption{The attempt-to-attempt mechanism. Conditional on being active ($C_{ij}=0$), the level context initializes the attempt, the terminal state yields completion $R_{ij}$ and signed margin $Q_{ij}$, completion updates expected experience, and experience and margin drive next-attempt churn. Churn is absorbing and gates whether attempt $j+1$ occurs under the same mechanisms and the same $K_i$. Within-attempt Markov steps are elided (dashed arrow).}
\label{fig:scm-attempt-dag}
\end{figure}

Figure~\ref{fig:scm-plate-dag} shows the full graph of the Match-3 SCM across the player, attempt, and gameplay scales. The oracle player state $K_i$ is shared across attempts and affects difficulty assignment, action selection, and the proxy measurements, but never the board transitions. Figure~\ref{fig:scm-attempt-dag} isolates the attempt-to-attempt mechanism. Churn is absorbing and gates whether the next attempt occurs, and the same mechanisms and the same $K_i$ generate every subsequent attempt.

\subsection{Engine Simulation Details}
\label{app:engine-details}

\begin{figure}[t]
\centering
\begin{subfigure}[t]{0.2\linewidth}
    \centering
    \includegraphics[
        width=\linewidth,
        height=0.23\textheight,
        keepaspectratio
    ]{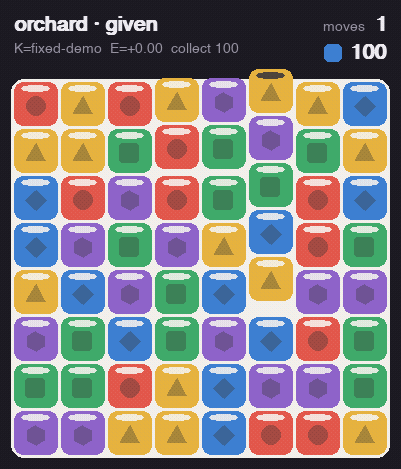}
    \caption{}
\end{subfigure}
\hspace{0.05\linewidth}
\begin{subfigure}[t]{0.2\linewidth}
    \centering
    \includegraphics[
        width=\linewidth,
        height=0.23\textheight,
        keepaspectratio
    ]{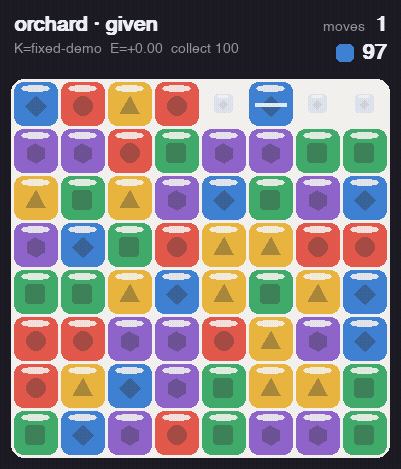}
    \caption{}
\end{subfigure}
\hspace{0.05\linewidth}
\begin{subfigure}[t]{0.2\linewidth}
    \centering
    \includegraphics[
        width=\linewidth,
        height=0.23\textheight,
        keepaspectratio
    ]{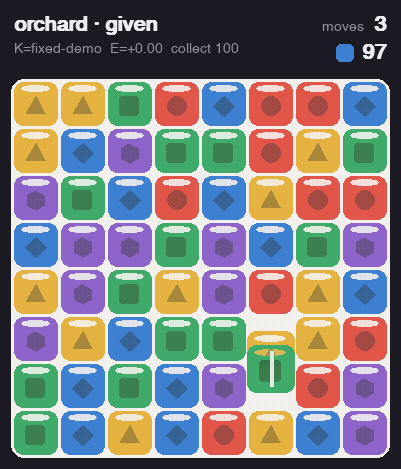}
    \caption{}
\end{subfigure}

\caption{Examples of gameplay mechanics: (a) a three-round cascade triggered by a single move, (b) creation and activation of a horizontal stripe, and (c) creation and activation of a vertical stripe.}
\label{fig:gameplay-examples}
\end{figure}

The engine resolves each legal swap to a settled state by applying matches, special-tile effects, cascades, gravity, and stochastic refill. Figure~\ref{fig:gameplay-examples} illustrates a representative cascade and the creation and activation of horizontal and vertical striped tiles.

\section{CaLA Architecture and Implementation Details} \label{app:cala-implementation}

\subsection{Overall Framework}

The CaLA framework consists of four interacting modules: 1) a variational history encoder $q_\eta$, 2) a player-policy Transformer $\pi_\psi$~\citep{vaswani2017attention}, 3) a transition world model $p_\theta$, and 4) a retention model $p_\gamma$. The history encoder compresses pretreatment behavior into a player representation. This representation conditions action selection and retention, while the transition model receives observable state, action, and task context.

The policy Transformer and world-model Transformer are separate networks. The dynamics model is first trained on natural-assignment transitions and then frozen. The history encoder, policy, retention model, and respective auxiliary heads are subsequently fitted jointly for each adjustment arm and assignment regime. Training therefore does not backpropagate through complete imagined trajectories. During inference, the posterior is computed before intervening on difficulty. Its sampled player contexts are held fixed throughout each imagined attempt and reused across candidate difficulties. Algorithm~\ref{alg:cala-rollout} gives the interventional rollout procedure.

\begin{algorithm}[t]
\caption{CaLA interventional rollout}
\label{alg:cala-rollout}
\begin{algorithmic}[1]
    \REQUIRE Pretreatment history $H_{i,<20}$, decision context $B_i$, difficulty $e$, posterior draws $S$, rollout replicates $N_{\mathrm{roll}}$.

    \STATE Infer the posterior $q_\eta(Z_i \mid H_{i,<20}, B_i)$ and draw $Z_i^{(s)}$.
    \STATE Set $E_i \leftarrow e$ and sample the opening board $S_{i0} \sim p_\theta(\cdot \mid L_i, D_i, e)$.

    \WHILE{trajectory not terminal ($t < T_i$)}
        \STATE Mask illegal actions on $S_{it}$ and sample $A_{it} \sim p_\psi(\cdot \mid S_{it}, Z_i^{(s)}, L_i, D_i, e)$.
        \STATE Advance the state $S_{i,t+1} \sim p_\theta(\cdot \mid S_{it}, A_{it}, L_i, D_i, e)$.
    \ENDWHILE

    \STATE Compute the trajectory summaries: completion $R_i(e)$, margin $Q_i(e)$ and updated experience $M_{i,21}(e)$.

    \STATE Evaluate churn risk $\widehat{\mu}_{\mathrm{WM},l}(e, B_i, Z_i^{(s)}) = p_\gamma(Y_i = 1 \mid R_i, Q_i, M_{i,21}, B_i, Z_i^{(s)}, L_i)$.

    \RETURN Average risk over the $S$ posterior draws and $N_{\mathrm{roll}}$ rollouts.
\end{algorithmic}
\end{algorithm}

The world-modeling choices are detailed in Appendix~\ref{app:world-model-architecture-and-rollout-evidence}.

\subsection{Variational History Encoder}

The player state is encoded using a variational history encoder with auxiliary likelihood heads. It is distinct from the latent representation used internally by LeWM and does not use an image-VAE board-reconstruction objective~\citep{kingma2014vae}. The considered prior is $p(z_i)=\mathcal{N}(0,I_4)$, while the posterior is a four-dimensional diagonal Gaussian:

\begin{align}
q_\eta(z_i\mid H_{i,<j}) 
&= \mathcal{N}\!\left(\mu_i,\operatorname{diag}\{\exp(2\ell_i)\} \right),
\label{eq:history-posterior}\\
z_i^{(s)} &= \mu_i+\exp(\ell_i)\odot\epsilon_i^{(s)},
\qquad \epsilon_i^{(s)}\sim\mathcal{N}(0,I_4).
\label{eq:latent-reparameterization}
\end{align}

Here $\mu_i$ and $\ell_i$ denote the encoder's predicted mean and log-standard deviation vectors, and $\ell_i$ is clamped to $[-6,3]$. Each sampled latent is concatenated with the observable base context:
\begin{equation}
w_i^{(s)}=[b_i,z_i^{(s)}]\in\mathbb{R}^{11}.
\label{eq:player-context}
\end{equation}

Overall, the encoder consists of four main stages:

\begin{enumerate}[label=\arabic*)]
    \item \textbf{Step encoding.} The width-$64$ color embeddings are averaged over the 64 board cells. This representation is concatenated with a width-$64$ action embedding and two counters, normalized by $22$ for moves and $32$ for goals. A $130\rightarrow64\rightarrow64$ GELU MLP produces the step embedding. Under this architecture, the history encoder does not directly encode spatial coordinates or special-kind grids.

    \item \textbf{Attempt encoding.} Valid step embeddings are mean-pooled and concatenated with width-$64$ level and tier embeddings, difficulty, completion, margin, and the $12$ other behavioral indicators. A $207\rightarrow64\rightarrow64$ GELU MLP produces one embedding per attempt.

    \item \textbf{History aggregation.} A single-layer GRU~\citep{cho2014gru} with input and hidden width $64$ processes the ordered historical attempts. Packed sequences exclude padding. Again, we note that only attempts preceding the target are included.
    
    \item \textbf{Posterior readout.} The final GRU state is concatenated with a learned width-$64$ constant and passed through a $128\rightarrow8$ linear head. The eight outputs specify the posterior parameters: four means and four log-standard deviations. The constant replaces unavailable baseline-evidence fields. It is not a measured player-specific feature. A separate learned vector handles empty histories.
\end{enumerate}

Overall, the encoder contains $65,096$ parameters. We set the KL weight for variational CaLA to $0.01$. The deterministic-$Z$ arm uses the encoder mean with a latent penalty of $10^{-3}\lVert z_i\rVert_2^2$.

\subsection{Player-Policy Transformer}

The player-policy Transformer predicts legal-swap probabilities from the current board, counters, and player context. It uses $67$ input tokens, of which $64$ represent tiles, with one additional token each for player context, remaining moves, and remaining goals. Tile tokens combine color, row, column, and goal-color indicator embeddings. The architecture consists of four pre-normalized Transformer blocks with width $128$, four attention heads, and a feed-forward width of $512$. It uses GELU activations and zero dropout. A final layer normalization is then applied.

For each candidate action, the policy concatenates the representations of its two endpoints and the context token with two immediate-action features. The first is the total number of clears in the round, and the second is the number of goal-color clears computed from the color board. These features are standardized across legal actions in the current state. A $386\rightarrow128\rightarrow1$ head, with a GELU activation function, produces the action score.

The action vocabulary contains $128$ slots in total, encoding two swap directions per cell, of which $112$ are the in-bounds adjacent swaps on an $8\times8$ board and the remaining $16$ denote out-of-board swaps that are never legal. State-dependent masks remove illegal actions before normalization. The policy architecture contains $850,817$ parameters for the $11$-dimensional CaLA context.

\section{World-Model Architecture and Rollout Evidence}
\label{app:world-model-architecture-and-rollout-evidence}

To understand how difficulty affects churn and player behavior, we must approximate the evolution of the symbolic Match-3 state under a selected action. The world model represents this dynamical system: each transition begins from a settled board and ends after the selected swap and all resulting game mechanics have been resolved. Stochastic refill requires the model both to predict state changes and to represent uncertainty over the resulting boards and goal progress. We therefore assess prediction accuracy at the individual-transition level and evaluate model behavior during closed-loop generation of its own states.

As explained previously, the input contains parallel $8\times8$ color and special-kind grids, the goal color, remaining moves, and remaining goals. Special kinds distinguish ordinary tiles from horizontal and vertical striped tiles. The selected action identifies the adjacent cells to swap, while the task context specifies the level, baseline tier, and served difficulty. The transition model uses these observable state and task variables without player-identity or player-history inputs.

Each model predicts the next color and special-kind grids, goal progress, and termination. The move counter is decremented deterministically. This common interface allows the architectures to be compared on the same observable prediction task, even though their internal representations differ.

\subsection{Architecture Families}
We consider three action-conditioned model families: an autoregressive Transformer~\citep{vaswani2017attention} and two models based on latent representations.

\begin{enumerate}[label=\arabic*)]
    \item \textbf{Next-Board Transformer.} This is a conventional autoregressive model. The model represents each board cell as a token containing color, special kind, row and column position, whether the tile matches the goal color, and its role in the selected swap. Additional tokens encode the action, task context, and counters. The Transformer encoder combines these inputs while retaining a representation for every board position. This architecture predicts observable state directly and does not maintain a recurrent latent state between transitions. Its training objective includes board likelihood together with the shared goal, cleared-mask, special-state, and terminal losses.

    \item \textbf{LeJEPA.} This is an action-conditioned variant of the Joint-Embedding Predictive Architecture (JEPA)~\citep{assran2023ijepa} that dispenses with an exponential-moving-average target encoder and uses the SIGReg regularizer introduced with LeJEPA~\citep{balestriero2025lejepa}. A spatial encoder maps the current state to per-cell and global latent representations. An action-conditioned Transformer predictor then predicts the latent representation of the next state. The target representation is obtained by encoding the observed next state with the online encoder and stopping gradients through that target branch. The representation objective minimizes the squared error between predicted and target latents. Without regularization, the architecture collapses to a trivial solution. SIGReg discourages such degenerate representations by comparing random projections of the embeddings with a standard Gaussian reference distribution. Observable readout heads produce board and goal predictions for evaluation. During multi-step prediction, LeJEPA propagates its predicted latent representation between transitions.

    \item \textbf{LeWM.} This architecture extends our adapted LeJEPA model by incorporating actions into an end-to-end latent world model. It uses an action-conditioned latent predictor with a shared target encoder. The latent prediction objective propagates gradients through both the current-state and next-state encoding branches. SIGReg is applied to both branches using flattened state representations. For every LeWM result reported here, the predicted color and special-kind grids are decoded and then re-encoded before the next transition is predicted. The model therefore repeatedly derives its input representation from its generated symbolic state. We use the name LeWM consistently for this rollout configuration at all three parameter bands.
\end{enumerate}

We additionally note that the latent-based representation models also supervise color information in their encoded states. Their board and special readouts receive detached predicted latents, whereas goal-related and terminal losses retain gradient paths to the predicted representation.

\subsection{Experimental Setup}

For each model family, we consider three common trainable-parameter targets: $1.0$M, $5.2$M, and $14.8$M. The parameter counts include the prediction heads.

All results are reported across three training seeds to reduce sensitivity to random initialization and other stochastic factors.

All models are trained for $11,400$ updates using AdamW~\citep{loshchilov2019adamw} with a learning rate of $3\times10^{-4}$, weight decay of $10^{-4}$, a batch size of $32$ episodes, and gradient clipping at $1$. No learning-rate schedule is used. Variable-length episodes are padded, and padding is excluded from the losses.

Checkpoints are selected by minimizing the sum of the validation cleared-mask and goal losses. The tables report the selected checkpoint within the common update budget.

\subsection{Evaluation Protocol}

We use three complementary evaluation modes to assess next-state prediction at different horizons and what we refer to as \emph{imagination}. These modes distinguish local prediction errors from errors that accumulate during rollout:

\begin{enumerate}[label=\arabic*)]
    \item \textbf{Logged-action replay.} The model starts from a logged state, receives the recorded action sequence, and advances using its own predicted states. Predictions at each horizon are compared with the corresponding logged states. This measures multi-step prediction under the recorded behavior. The primary skill results use argmax decoding. For this specific setting, the recorded actions are not guaranteed to remain legal after the predicted board diverges, so replay is reported separately from autonomous imagination.
    \item \textbf{Engine-conditioned prediction.} At every step, the engine samples a uniformly random legal action from its current state and the model receives that state-action pair. The model predicts one transition and is then reset to the next true state. This isolates local transition error from errors caused by repeatedly feeding back model predictions. A score reported at step seven in this mode remains an engine-conditioned one-step score evaluated later along the trajectory.
    \item \textbf{Autonomous imagination.} The model and engine begin from the same validation anchor, and the engine remains the sole action source. At every step, it samples a uniformly random legal action from its current board. The identical action is supplied to the model, which advances from its own previously predicted board, special state, and goal counter. This paired-action construction holds the action sequence fixed across the two systems while allowing model-state errors to compound over the rollout. Once the boards diverge, a sampled action can be illegal on the model's predicted board. The model does not check legality and predicts the next transition from any supplied action, so these mismatches surface as state error rather than as a protocol exception. Terminal states are retained when forming subsequent horizon-level summaries.
\end{enumerate}

\subsection{Metrics}

Our primary board-prediction metric is \emph{persistence-normalized skill}. The persistence baseline predicts that the board remains unchanged: at every horizon, each cell retains its color from the initial observed board. Thus, if $C_t$ denotes the initial color grid, the baseline prediction is $\widehat{C}^{\mathrm{pers}}_{t+k}=C_t$ for every $k$. This baseline is informative because many cells can retain their colors between observations, allowing an unchanged-board prediction to achieve relatively high raw accuracy.

Let $\mathrm{Err}_{\mathrm{model},k}$ and $\mathrm{Err}_{\mathrm{pers},k}$ denote the fractions of incorrectly predicted cell colors for the model and persistence baseline, respectively, aggregated over the evaluated boards and cells. We define:

\begin{equation}
\mathrm{Skill}_k
=
1-
\frac{\mathrm{Err}_{\mathrm{model},k}}
     {\mathrm{Err}_{\mathrm{pers},k}}
=
\frac{\mathrm{Acc}_{\mathrm{model},k}
      -\mathrm{Acc}_{\mathrm{pers},k}}
     {1-\mathrm{Acc}_{\mathrm{pers},k}}.
\label{eq:persistence-skill}
\end{equation}

We note that a score of zero matches persistence, a positive score indicates an improvement over the baseline, and a negative score indicates worse performance. A score of one corresponds to perfect color prediction. For instance, an overall skill score of $0.30$ represents a $30\%$ reduction in color-prediction error relative to persistence. We consider three horizons, $k \in \{1, 3, 7\}$.

Additionally, for any given initial state and horizon, let $\{X_i\}_{i=1}^{M}$ be the model-generated samples and $\{Y_j\}_{j=1}^{N}$ be the engine-generated reference samples. The empirical energy score~\citep{gneiting2007scoring} is:

\begin{equation}
\widehat{\mathrm{ES}}
=
\frac{1}{MN}
\sum_{i=1}^{M}\sum_{j=1}^{N} d(X_i,Y_j)
-
\frac{1}{2M(M-1)}
\sum_{\substack{i,i'=1\\i\neq i'}}^{M}
d(X_i,X_{i'}),
\label{eq:energy-score}
\end{equation}

where $M>1$ and $d(\cdot,\cdot)$ is the per-cell mismatch count between sampled boards for the board score and the absolute difference of remaining goal counts for the goal score. This quantity is computed for each initial state and then averaged across evaluated states.

\subsection{Experimental Results and Comparison}

Table~\ref{tab:results_WM} evaluates the Transformer, LeJEPA, and LeWM across three parameter settings, using persistence-normalized skill for logged-action replay and board and goal energy scores for autonomous stochastic imagination. Across all parameter capacities and prediction horizons, the Next-Board Transformer performs best. It is also parameter-efficient: at $5.2$M parameters, it outperforms the $14.8$M variants of both latent models across all replay horizons. At the $1.0$M scale, the latent models struggle, with LeWM performing near the persistence baseline and LeJEPA dropping below the baseline at longer horizons.

Under autonomous imagination, the Transformer obtains the lowest board energy scores overall, although LeWM scales competitively at higher capacities, approaching the best board energy at $14.8$M ($0.295$ versus $0.291$ at horizon 7) and achieving comparable one-step goal energy at $5.2$M ($0.629$ versus $0.630$).

\begin{figure}[t]
\centering
\includegraphics[width=\linewidth,keepaspectratio]{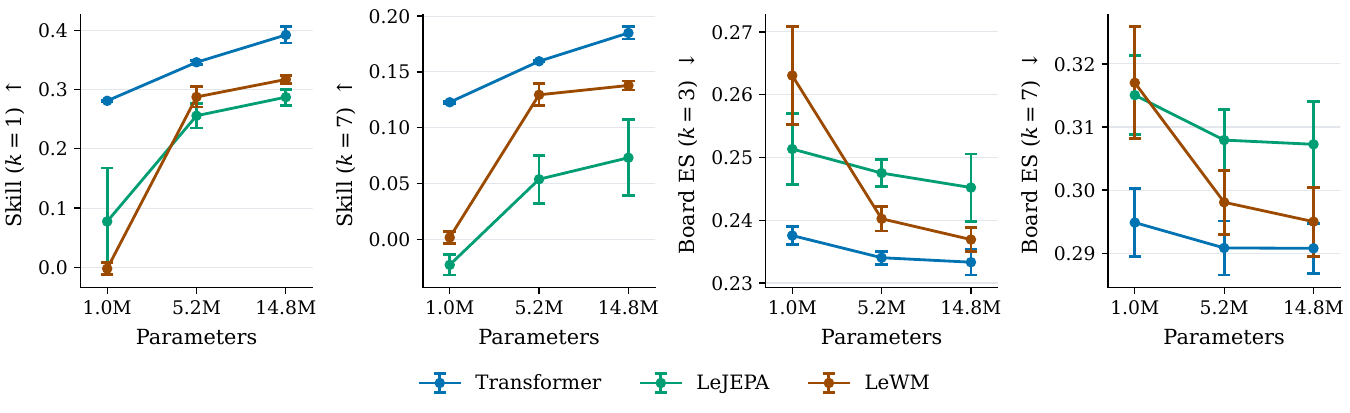}
\caption{Effect of model scale on selected world-model evaluation metrics.}
\label{fig:scaling_WM}
\end{figure}

Figure~\ref{fig:scaling_WM} illustrates the effect of model scale on selected metrics. Overall, the Transformer provides the best accuracy and rollout stability under the evaluated protocol, with LeWM emerging as the strongest latent alternative at larger capacities. We therefore focus the causal experiments on these two families. All evaluated checkpoints and logs are provided in the supplementary materials.

\begin{table}[t]
\caption{Performance across parameter scales. Skill is evaluated by logged-action replay with argmax color decoding and is normalized against an unchanged-board persistence baseline. Board and goal energy scores (ES) evaluate autonomous stochastic imagination against engine-generated reference trajectories. LeWM uses decoded-and-re-encoded states during rollout. We report means $\pm$ sample SDs across three training seeds. Higher skill and lower ES are better; bold indicates the best displayed mean for each metric.}
\label{tab:results_WM}
\centering
\scriptsize
\resizebox{\linewidth}{!}{%
\begin{tabular}{llcccccc}
\toprule
\textbf{World model} & \textbf{Parameters}
& \textbf{Skill $k_1$} $\uparrow$
& \textbf{Skill $k_3$} $\uparrow$
& \textbf{Skill $k_7$} $\uparrow$
& \textbf{Board ES $k_3$} $\downarrow$
& \textbf{Board ES $k_7$} $\downarrow$
& \textbf{Goal ES $k_1$} $\downarrow$ \\
\midrule
Transformer & 1.0M
& $0.281 \pm 0.001$ & $0.212 \pm 0.001$ & $0.123 \pm 0.001$
& $0.238 \pm 0.001$ & $0.295 \pm 0.005$ & $0.642 \pm 0.051$ \\
LeJEPA & 1.0M
& $0.077 \pm 0.090$ & $0.021 \pm 0.059$ & $-0.023 \pm 0.009$
& $0.251 \pm 0.006$ & $0.315 \pm 0.006$ & $0.665 \pm 0.058$ \\
LeWM & 1.0M
& $-0.002 \pm 0.010$ & $0.000 \pm 0.008$ & $0.001 \pm 0.005$
& $0.263 \pm 0.008$ & $0.317 \pm 0.009$ & $0.671 \pm 0.048$ \\
\midrule
Transformer & 5.2M
& $0.346 \pm 0.003$ & $0.269 \pm 0.001$ & $0.160 \pm 0.001$
& $0.234 \pm 0.001$ & $\mathbf{0.291 \pm 0.004}$ & $0.630 \pm 0.052$ \\
LeJEPA & 5.2M
& $0.256 \pm 0.021$ & $0.156 \pm 0.009$ & $0.054 \pm 0.021$
& $0.248 \pm 0.002$ & $0.308 \pm 0.005$ & $0.639 \pm 0.056$ \\
LeWM & 5.2M
& $0.288 \pm 0.017$ & $0.218 \pm 0.016$ & $0.129 \pm 0.010$
& $0.240 \pm 0.002$ & $0.298 \pm 0.005$ & $0.629 \pm 0.051$ \\
\midrule
Transformer & 14.8M
& $\mathbf{0.392 \pm 0.014}$ & $\mathbf{0.308 \pm 0.011}$ & $\mathbf{0.185 \pm 0.005}$
& $\mathbf{0.233 \pm 0.002}$ & $\mathbf{0.291 \pm 0.004}$ & $\mathbf{0.622 \pm 0.054}$ \\
LeJEPA & 14.8M
& $0.287 \pm 0.013$ & $0.186 \pm 0.028$ & $0.073 \pm 0.034$
& $0.245 \pm 0.005$ & $0.307 \pm 0.007$ & $0.631 \pm 0.057$ \\
LeWM & 14.8M
& $0.317 \pm 0.007$ & $0.238 \pm 0.006$ & $0.138 \pm 0.004$
& $0.237 \pm 0.002$ & $0.295 \pm 0.005$ & $0.626 \pm 0.046$ \\
\bottomrule
\end{tabular}%
}
\end{table}

\section{Additional Results}\label{app:additional-results}

\subsection{Direct-Adjustment Diagnostics}

Table~\ref{tab:direct-dr-aggregate} reports the aggregate Direct-DR results under both assignment regimes, and Table~\ref{tab:direct-dr-natural-level} reports the per-level results under natural assignment, including whole-player bootstrap uncertainty and treatment-support diagnostics. The full-sample optimum can have a low point regret while remaining unstable under resampling.

\begin{table}[t]
\caption{Aggregate Direct-DR performance against engine reference. Values are means $\pm$ sample SDs across three seeds. Lower is better.}
\label{tab:direct-dr-aggregate}
\centering
\footnotesize
\resizebox{0.8\linewidth}{!}{%
\begin{tabular}{llccc}
\toprule
\textbf{Regime} & \textbf{Adjustment arm}
& \shortstack{\textbf{Weighted curve}\\\textbf{MAE} $\downarrow$}
& \shortstack{\textbf{Mean population}\\\textbf{regret} $\downarrow$}
& \shortstack{\textbf{Pooled individual}\\\textbf{regret} $\downarrow$} \\
\midrule
Natural & No history
& $0.053 \pm 0.000$ & $0.059 \pm 0.002$ & $0.118 \pm 0.003$ \\
Natural & Handcrafted history
& $0.040 \pm 0.002$ & $0.026 \pm 0.012$ & $0.067 \pm 0.009$ \\
Natural & Deterministic $Z$
& $0.030 \pm 0.005$ & $0.010 \pm 0.007$ & $0.051 \pm 0.003$ \\
Natural & Variational CaLA
& $0.027 \pm 0.006$
& $\mathbf{0.007 \pm 0.000}$
& $\mathbf{0.045 \pm 0.005}$ \\
Natural & Oracle $K$
& $\mathbf{0.024 \pm 0.001}$
& $0.011 \pm 0.006$
& $0.049 \pm 0.002$ \\
\midrule
Randomized & No history
& $0.064 \pm 0.001$ & $0.022 \pm 0.001$ & $0.065 \pm 0.002$ \\
Randomized & Handcrafted history
& $0.073 \pm 0.002$ & $0.043 \pm 0.000$
& $\mathbf{0.043 \pm 0.007}$ \\
Randomized & Deterministic $Z$
& $\mathbf{0.058 \pm 0.003}$
& $0.024 \pm 0.004$
& $0.045 \pm 0.003$ \\
Randomized & Variational CaLA
& $0.064 \pm 0.004$
& $\mathbf{0.020 \pm 0.010}$
& $0.049 \pm 0.003$ \\
Randomized & Oracle $K$
& $0.066 \pm 0.003$ & $0.024 \pm 0.001$ & $0.053 \pm 0.003$ \\
\bottomrule
\end{tabular}%
}
\end{table}

\begin{table}[p]
\caption{Per-level Direct-DR performance under natural assignment against the engine reference. Point estimates use the mean curve across three runs. Bootstrap 95\% ranges use 2,000 joint player resamples and are reported as midpoint $\pm$ half-width. Selection probability measures selection of the engine optimum. Support is the fraction of player--difficulty pairs whose fitted propensity density exceeds the floor $g_{\min} = 0.02$.}
\label{tab:direct-dr-natural-level}
\centering
\scriptsize
\resizebox{\linewidth}{!}{%
\begin{tabular}{llccccccc}
\toprule
\textbf{Level} & \textbf{Adjustment arm}
& \textbf{Curve MAE} $\downarrow$
& \shortstack{\textbf{Selected}\\\textbf{difficulty}}
& \shortstack{\textbf{Engine}\\\textbf{optimum}}
& \shortstack{\textbf{Point}\\\textbf{regret} $\downarrow$}
& \shortstack{\textbf{Bootstrap regret}\\\textbf{95\% range}}
& \shortstack{\textbf{Probability of}\\\textbf{selecting $e^*$} $\uparrow$}
& \textbf{Support} $\uparrow$ \\
\midrule
Orchard & No history
& 0.050 & $+2.00$ & $+0.25$ & 0.058
& $0.039 \pm 0.019$ & 0.3\% & \textbf{100.0\%} \\
Orchard & Handcrafted history
& 0.030 & $+1.50$ & $+0.25$ & 0.040
& $0.056 \pm 0.056$ & 4.6\% & 64.9\% \\
Orchard & Deterministic $Z$
& 0.021 & $+0.25$ & $+0.25$ & \textbf{0.000}
& $0.029 \pm 0.029$ & 35.0\% & 65.1\% \\
Orchard & Variational CaLA
& 0.015 & $+0.25$ & $+0.25$ & \textbf{0.000}
& $0.018 \pm 0.018$ & \textbf{52.6\%} & 65.1\% \\
Orchard & Oracle $K$
& \textbf{0.013} & $+0.25$ & $+0.25$ & \textbf{0.000}
& $0.032 \pm 0.032$ & 46.0\% & 64.4\% \\
\midrule
Harbour & No history
& 0.047 & $-1.75$ & $-0.75$ & 0.053
& $0.063 \pm 0.061$ & 1.4\% & \textbf{100.0\%} \\
Harbour & Handcrafted history
& 0.052 & $-1.50$ & $-0.75$ & 0.021
& $0.037 \pm 0.028$ & 0.4\% & 65.0\% \\
Harbour & Deterministic $Z$
& \textbf{0.033} & $-1.25$ & $-0.75$ & \textbf{0.013}
& $0.033 \pm 0.033$ & \textbf{4.1\%} & 67.6\% \\
Harbour & Variational CaLA
& 0.035 & $-1.50$ & $-0.75$ & 0.021
& $0.033 \pm 0.033$ & 2.9\% & 68.6\% \\
Harbour & Oracle $K$
& 0.034 & $-1.50$ & $-0.75$ & 0.021
& $0.034 \pm 0.032$ & 2.1\% & 65.3\% \\
\midrule
Foundry & No history
& 0.069 & $+2.00$ & $+0.50$ & 0.056
& $0.037 \pm 0.019$ & 0.3\% & \textbf{100.0\%} \\
Foundry & Handcrafted history
& 0.045 & $+1.00$ & $+0.50$ & 0.004
& $0.085 \pm 0.085$ & 14.3\% & 68.0\% \\
Foundry & Deterministic $Z$
& 0.048 & $+0.25$ & $+0.50$ & 0.005
& $0.030 \pm 0.030$ & 13.5\% & 69.9\% \\
Foundry & Variational CaLA
& \textbf{0.032} & $+0.50$ & $+0.50$ & \textbf{0.000}
& $0.030 \pm 0.030$ & 23.9\% & 69.7\% \\
Foundry & Oracle $K$
& 0.034 & $+0.75$ & $+0.50$ & 0.002
& $0.024 \pm 0.024$ & \textbf{29.7\%} & 69.4\% \\
\bottomrule
\end{tabular}%
}
\end{table}

\begin{figure}[t]
\centering
\includegraphics[width=\textwidth]{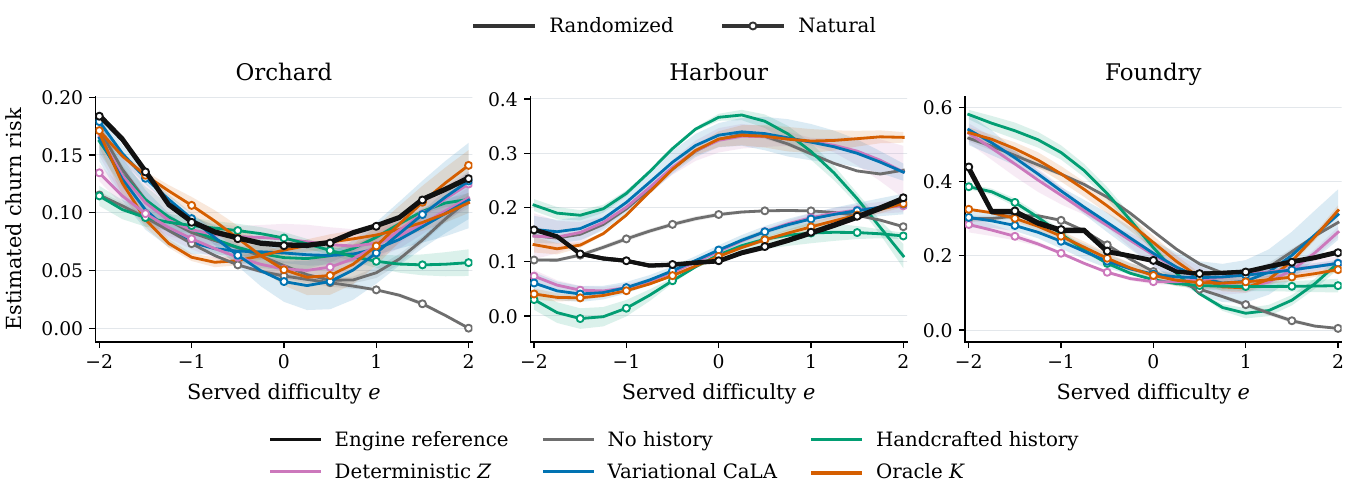}
\caption{Direct-DR population intervention curves for held-out players under randomized (solid) and natural (hollow markers) assignment. Black curves show the paired engine reference. Colored curves and shaded bands show means and $\pm$ one sample SD across three seeds.}
\label{fig:direct-adjustment-population}
\end{figure}

\begin{figure}[t]
\centering
\includegraphics[width=\linewidth]{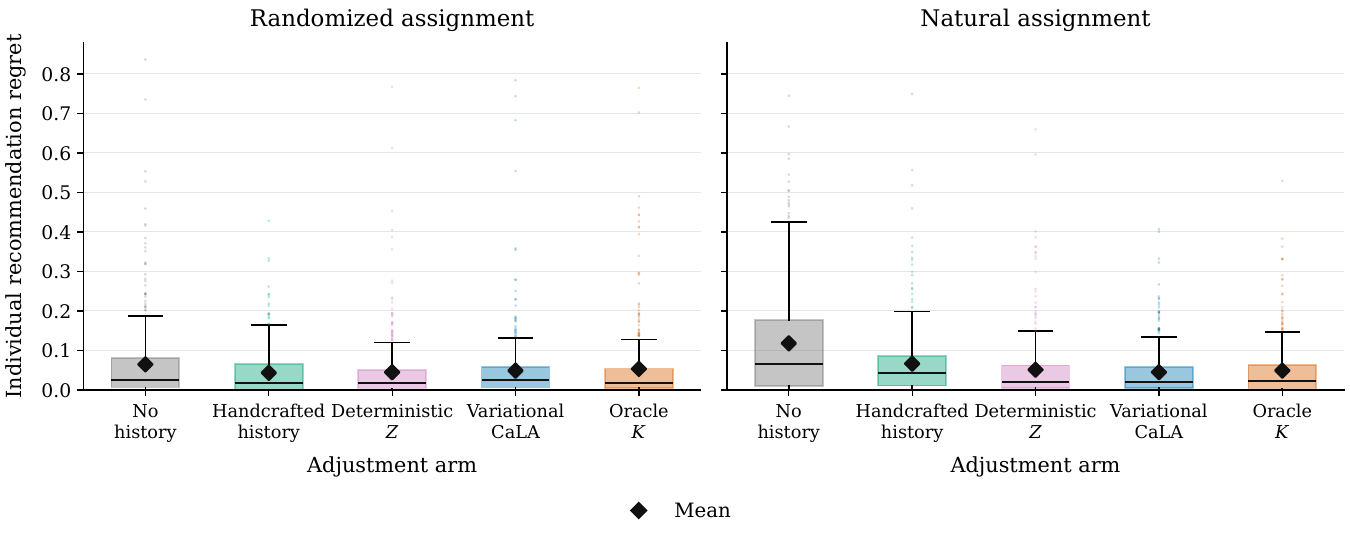}
\caption{Direct-DR individual recommendation regret under randomized and natural assignment, pooled across the three level families. Each player's regret is averaged across three seeds. Boxes summarize supported individual risks, and diamonds mark means.}
\label{fig:direct-adjustment-individual}
\end{figure}

Figure~\ref{fig:direct-adjustment-population} shows the Direct-DR population curves against the paired engine interventions, and Figure~\ref{fig:direct-adjustment-individual} shows the individual-regret distributions. All history-conditioned arms improve on the no-history baseline in aggregate, although oracle $K$ is not a lower bound because every arm still depends on estimated nuisance functions.

\subsection{Per-Level WM-MC Results}

Table~\ref{tab:wm-mc-natural-level} expands the natural-assignment WM-MC aggregate in Table~\ref{tab:wm-aggregate}. Variational CaLA has the lowest aggregate mean curve error among non-oracle arms, but the seed-specific recommendations show that low curve error does not imply exact recovery of the engine optimum.

\begin{table}[p]
\caption{WM-MC population performance under natural assignment
against the engine reference. Values are means $\pm$ sample SDs across three seeds. Recommendations follow the considered seed order (4201, 4202, 4203). Bold compares the latter rows only.}
\label{tab:wm-mc-natural-level}
\centering
\scriptsize
\resizebox{\linewidth}{!}{%
\begin{tabular}{llcccc}
\toprule
\textbf{Level} & \textbf{Adjustment arm}
& \textbf{Curve MAE} $\downarrow$
& \shortstack{\textbf{Selected}\\\textbf{difficulty}}
& \shortstack{\textbf{Engine}\\\textbf{optimum}}
& \shortstack{\textbf{Point}\\\textbf{regret} $\downarrow$} \\
\midrule
Orchard & No history
& $0.066 \pm 0.004$
& $(+1.75,+1.75,+1.25)$ & $+0.25$
& $0.040 \pm 0.014$ \\
Orchard & Handcrafted history
& $0.018 \pm 0.001$
& $(+0.25,-0.25,-0.50)$ & $+0.25$
& $0.003 \pm 0.003$ \\
Orchard & Deterministic $Z$
& $0.020 \pm 0.007$
& $(-0.25,+0.00,+0.25)$ & $+0.25$
& $0.001 \pm 0.001$ \\
Orchard & Variational CaLA
& $\mathbf{0.016 \pm 0.007}$
& $(+0.25,+0.00,+0.25)$ & $+0.25$
& $\mathbf{0.000 \pm 0.000}$ \\
Orchard & Oracle $K$
& $0.007 \pm 0.001$
& $(+0.25,+0.25,+0.25)$ & $+0.25$
& $0.000 \pm 0.000$ \\
\midrule
Harbour & No history
& $0.093 \pm 0.006$
& $(+0.25,+1.00,+1.00)$ & $-0.75$
& $0.049 \pm 0.021$ \\
Harbour & Handcrafted history
& $\mathbf{0.013 \pm 0.001}$
& $(-0.75,-0.50,-1.00)$ & $-0.75$
& $\mathbf{0.003 \pm 0.005}$ \\
Harbour & Deterministic $Z$
& $0.019 \pm 0.004$
& $(-0.50,+0.00,-0.50)$ & $-0.75$
& $0.004 \pm 0.004$ \\
Harbour & Variational CaLA
& $0.023 \pm 0.001$
& $(-0.25,-0.25,-0.25)$ & $-0.75$
& $0.006 \pm 0.000$ \\
Harbour & Oracle $K$
& $0.021 \pm 0.003$
& $(-0.50,-0.25,+0.00)$ & $-0.75$
& $0.005 \pm 0.003$ \\
\midrule
Foundry & No history
& $0.093 \pm 0.005$
& $(+1.00,+1.25,+1.75)$ & $+0.50$
& $0.021 \pm 0.019$ \\
Foundry & Handcrafted history
& $0.028 \pm 0.009$
& $(+0.50,+0.25,+0.25)$ & $+0.50$
& $0.004 \pm 0.003$ \\
Foundry & Deterministic $Z$
& $0.029 \pm 0.003$
& $(+0.25,+0.50,+0.50)$ & $+0.50$
& $0.002 \pm 0.003$ \\
Foundry & Variational CaLA
& $\mathbf{0.024 \pm 0.003}$
& $(+0.75,+0.50,+0.50)$ & $+0.50$
& $\mathbf{0.001 \pm 0.001}$ \\
Foundry & Oracle $K$
& $0.024 \pm 0.007$
& $(+0.25,+0.25,+0.75)$ & $+0.50$
& $0.004 \pm 0.002$ \\
\bottomrule
\end{tabular}%
}
\end{table}

\begin{figure}[p]
\centering
\includegraphics[width=\linewidth]{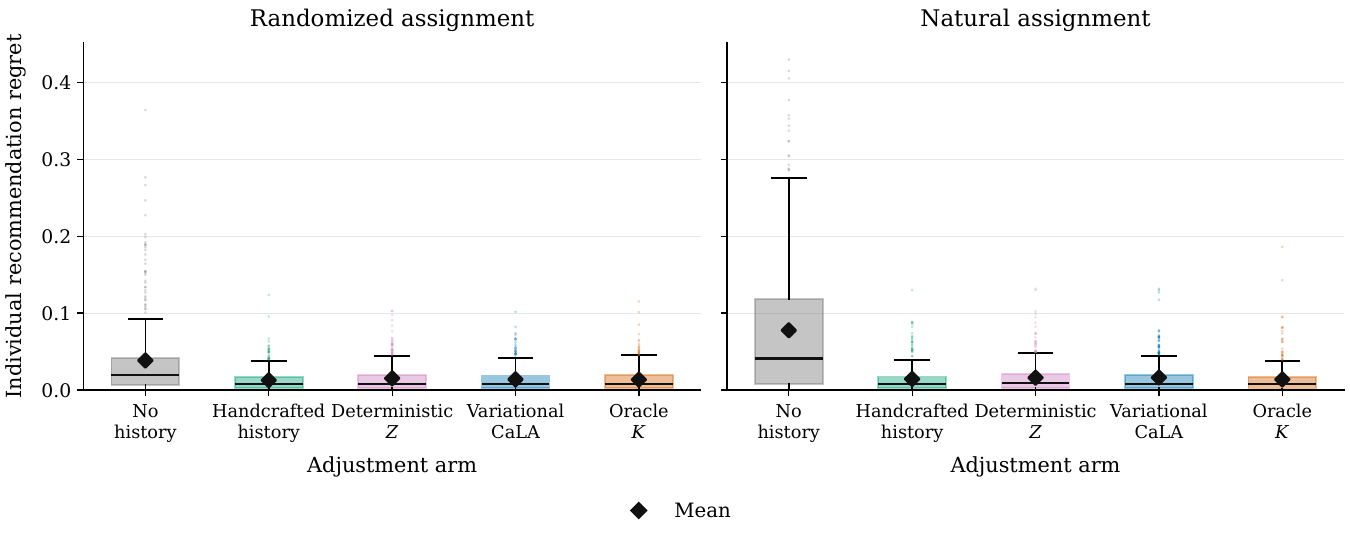}
\caption{WM-MC individual recommendation regret under randomized and natural assignment, pooled across the three level families. Each player's regret is averaged across three seeds. Boxes summarize supported individual risks, and diamonds mark means.}
\label{fig:wm-individual-regret}
\end{figure}

\subsection{World-Model Doubly Robust Evaluation}

\begin{figure}[t]
\centering
\includegraphics[width=\textwidth]{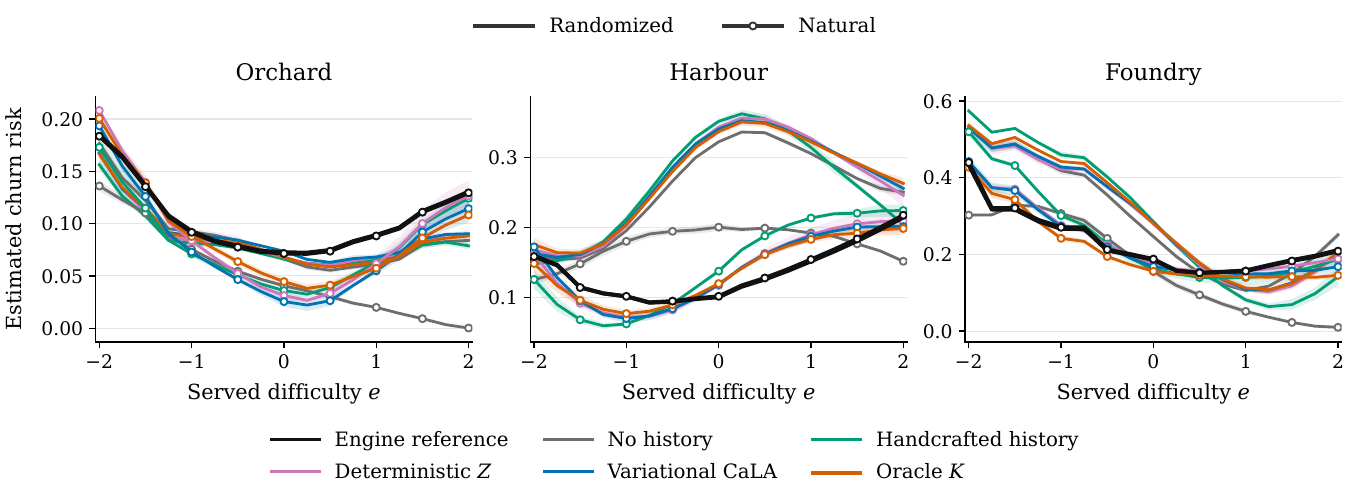}
\caption{WM-DR population intervention curves under randomized (solid) and natural (hollow markers) assignment. Black curves show the engine reference. Colored curves and shaded bands show the mean and $\pm$ one sample SD across three seeds.}
\label{fig:wm-dr-appendix}
\end{figure}

We evaluate whether augmenting world-model predictions with observed-outcome residuals improves population intervention estimation. WM-DR uses the same world-model predictions, adjustment contexts, test players, difficulty grid, and engine reference as WM-MC, with results reported as means and sample SDs across three seeds. Table~\ref{tab:wm-dr-population} reports aggregate population performance, and Figure~\ref{fig:wm-dr-appendix} plots the intervention curves.

With natural assignment, WM-DR performs substantially better when incorporating player history than the no-history baseline. Variational CaLA records the lowest weighted curve MAE among all adjustment contexts, reducing curve error by roughly $58\%$ relative to that baseline. CaLA also markedly reduces population regret, although deterministic $Z$ obtains the lowest regret under WM-DR. As illustrated in Figure~\ref{fig:wm-dr-appendix}, the history-conditioned estimates more closely track the engine intervention curves, supporting the value of pretreatment player context for population response estimation under natural assignment.

Under randomized assignment, the curve-error deterioration relative to WM-MC appears across all adjustment contexts, not only oracle $K$.

\begin{table}[t]
\caption{WM-DR population performance against the engine reference. Values are means $\pm$ sample SDs across three seeds. Bold compares the latter rows only.}
\label{tab:wm-dr-population}
\centering
\scriptsize
\resizebox{\linewidth}{!}{%
\begin{tabular}{llcc}
\toprule
\textbf{Regime} & \textbf{Adjustment arm}
& \textbf{Weighted curve MAE} $\downarrow$
& \textbf{Mean population regret} $\downarrow$ \\
\midrule
Natural & No history
& $0.058 \pm 0.000$ & $0.060 \pm 0.000$ \\
Natural & Handcrafted history
& $0.030 \pm 0.004$ & $0.006 \pm 0.003$ \\
Natural & Deterministic $Z$
& $\mathbf{0.020 \pm 0.001}$ & $\mathbf{0.005 \pm 0.003}$ \\
Natural & Variational CaLA
& $0.023 \pm 0.001$ & $0.007 \pm 0.003$ \\
Natural & Oracle $K$
& $0.021 \pm 0.001$ & $0.011 \pm 0.006$ \\
\midrule
Randomized & No history
& $0.061 \pm 0.001$ & $0.021 \pm 0.000$ \\
Randomized & Handcrafted history
& $0.074 \pm 0.001$ & $0.032 \pm 0.002$ \\
Randomized & Deterministic $Z$
& $0.067 \pm 0.002$ & $\mathbf{0.019 \pm 0.006}$ \\
Randomized & Variational CaLA
& $\mathbf{0.066 \pm 0.001}$ & $0.026 \pm 0.000$ \\
Randomized & Oracle $K$
& $0.068 \pm 0.002$ & $0.019 \pm 0.006$ \\
\bottomrule
\end{tabular}%
}
\end{table}

\subsection{Sensitivity to World-Model Architecture and Capacity}

The main experiments use a Transformer with a target parameter count of $14.8$M as the default world model. Here, we examine whether CaLA's performance extends to a smaller Transformer and to LeWM, whose architecture and predictive evaluation are described in Appendix~\ref{app:world-model-architecture-and-rollout-evidence}. We evaluate both families at parameter targets of $5.2$M and $14.8$M. All configurations use the same three seeds, corrected engine reference, and evaluation protocol.

\begin{figure}[t]
\centering
\includegraphics[width=\textwidth]{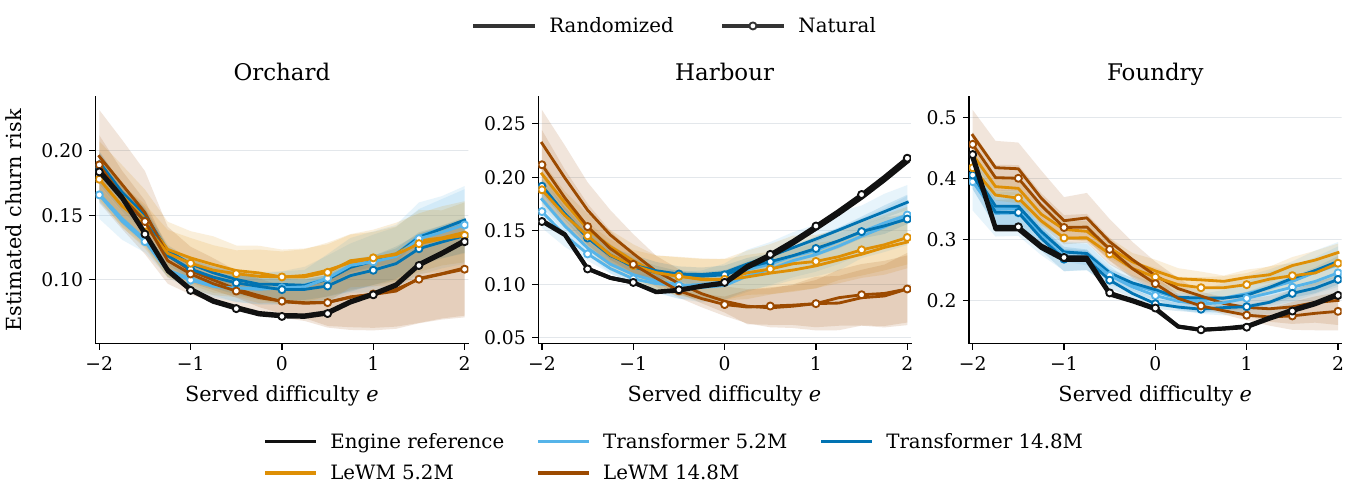}
\caption{WM-MC intervention curves with variational CaLA across world-model architectures and parameter scales, under randomized (solid) and natural (hollow markers) assignment. Black curves denote the engine reference. Colored curves and bands show means and $\pm$ one sample SD across three seeds.}
\label{fig:architecture-wm-mc}
\end{figure}

Figure~\ref{fig:architecture-wm-mc} compares WM-MC response curves across the four world models while holding the adjustment context fixed to variational CaLA. Table~\ref{tab:cala-wm-capacity} quantifies the curve errors and decision regrets. The lowest mean population regret belongs to LeWM at $5.2$M under both assignment regimes. Increasing capacity does not consistently improve downstream performance.

\begin{table}[t]
\caption{WM-MC performance with variational CaLA across world-model architectures against the engine reference. Historical current-policy predictions use 128 imagined trajectories per player--difficulty.
Values are means $\pm$ sample SDs across three seeds.}
\label{tab:cala-wm-capacity}
\centering
\scriptsize
\resizebox{\linewidth}{!}{%
\begin{tabular}{lllccc}
\toprule
\textbf{Regime} & \textbf{World model} & \textbf{Parameters}
& \shortstack{\textbf{Weighted curve}\\\textbf{MAE} $\downarrow$}
& \shortstack{\textbf{Mean population}\\\textbf{regret} $\downarrow$}
& \shortstack{\textbf{Pooled individual}\\\textbf{regret} $\downarrow$} \\
\midrule
Natural & Transformer & 5.2M
& $0.022 \pm 0.008$
& $0.009 \pm 0.008$
& $\mathbf{0.017 \pm 0.001}$ \\
Natural & Transformer & 14.8M
& $\mathbf{0.021 \pm 0.005}$
& $0.006 \pm 0.004$
& $0.017 \pm 0.002$ \\
Natural & LeWM & 5.2M
& $0.031 \pm 0.009$
& $\mathbf{0.002 \pm 0.001}$
& $0.019 \pm 0.001$ \\
Natural & LeWM & 14.8M
& $0.032 \pm 0.006$
& $0.019 \pm 0.013$
& $0.022 \pm 0.003$ \\
\midrule
Randomized & Transformer & 5.2M
& $0.022 \pm 0.007$
& $0.005 \pm 0.002$
& $\mathbf{0.015 \pm 0.001}$ \\
Randomized & Transformer & 14.8M
& $\mathbf{0.022 \pm 0.002}$
& $0.007 \pm 0.003$
& $0.016 \pm 0.002$ \\
Randomized & LeWM & 5.2M
& $0.034 \pm 0.011$
& $\mathbf{0.004 \pm 0.000}$
& $0.017 \pm 0.002$ \\
Randomized & LeWM & 14.8M
& $0.037 \pm 0.012$
& $0.024 \pm 0.022$
& $0.025 \pm 0.008$ \\
\bottomrule
\end{tabular}%
}
\end{table}

\begin{figure}[t]
\centering
\includegraphics[width=\linewidth]{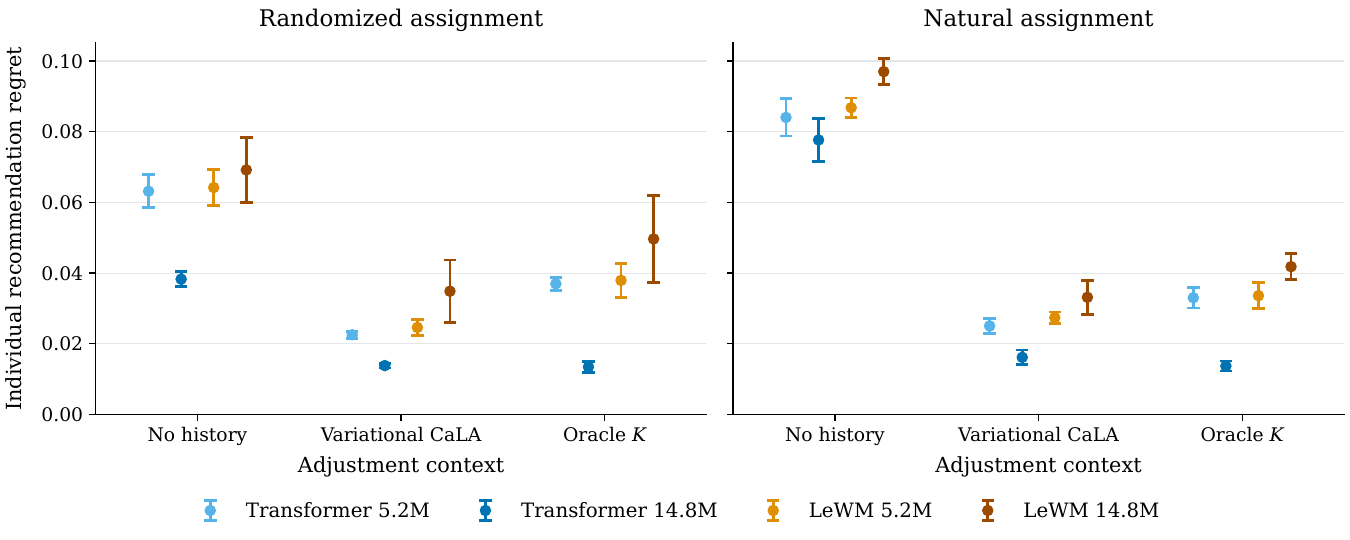}
\caption{Individual recommendation regret across world models and the three adjustment contexts evaluated for every model. Points show mean pooled individual regret across three seeds. Error bars indicate one sample SD.}
\label{fig:architecture-individual-regret}
\end{figure}

Figure~\ref{fig:architecture-individual-regret} compares the three adjustment contexts evaluated for every world model: no history, variational CaLA, and oracle $K$. CaLA has the smallest pooled individual regret among these contexts for each world model in both assignment regimes, so its personalization benefit extends across architecture families and parameter scales. Among the four CaLA configurations, the $5.2$M Transformer attains the lowest pooled individual regret under randomized assignment and ties the $14.8$M Transformer under natural assignment (Table~\ref{tab:cala-wm-capacity}). These comparisons are descriptive summaries across three seeds.

\end{document}